\pdfoutput=1

\documentclass[11pt]{article}

\usepackage{acl}

\usepackage{times}
\usepackage{latexsym}

\usepackage[T1]{fontenc}

\usepackage[utf8]{inputenc}

\usepackage{microtype}
\usepackage{graphicx}
\usepackage{colortbl}

\usepackage{amsmath}
\usepackage{float}
\usepackage{times}
\usepackage{latexsym}
\usepackage[T1]{fontenc}
\usepackage{soul}
\usepackage{multirow}
\usepackage{makecell}
\usepackage{boxedminipage}
\usepackage[utf8]{inputenc}
\usepackage{microtype}
\usepackage{graphicx}
\usepackage{soul}
\usepackage{float}
\usepackage{comment}
\usepackage{adjustbox}
\usepackage{booktabs}
\usepackage{amsmath}
\usepackage{mathabx}
\usepackage[shortlabels]{enumitem}
\usepackage[normalem]{ulem}
\usepackage{xspace}
\usepackage{array}
\usepackage{framed}
\usepackage{subcaption}
\usepackage{multicol}
\usepackage{xcolor}
\title{Issues with AUC and Alternative Metrics}
\title{Investigating the AUC Metric and Proposing Alternatives}
\title{Investigating Issues with the Area Under the Curve Metric and Proposing Solutions}
\title{Investigating the Failure Modes of the AUC metric and Exploring Alternatives for Evaluating Systems in Safety Critical Applications}

\author{Swaroop Mishra $\;$ Anjana Arunkumar $\;$ Chitta Baral
\\\\
 Arizona State University }
\begin{document}
\maketitle
\begin{abstract}
With the increasing importance of safety requirements associated with the use of black box models, evaluation of selective answering capability of models has been critical. Area under the curve (AUC) is used as a metric for this purpose. We find limitations in AUC; e.g., a model having higher AUC is not always better in performing selective answering. We propose three alternate metrics that fix the identified limitations. On experimenting with ten models, our results using the new metrics show that \textit{newer and larger pre-trained models do not necessarily show better performance in selective answering}. We hope our insights will help develop better models tailored for safety-critical applications.
\end{abstract}

\section{Introduction}
Humans have the capability to gauge how much they know; this leads them to abstain from answering whenever they are not confident about an answer. Such a  selective answering capability \cite{kamath2020selective, varshney2020s, garg2021will} is also essential for machine learning systems, especially in the case of safety critical applications like healthcare, where incorrect answering can result in critically negative consequences.\\ 

Area under the curve (AUC) has been used as a metric to evaluate models based on their selective answering capability. AUC involves finding model coverage and accuracy, at various confidence values. MaxProb-- maximum softmax probability of models' predictions-- has been used as a strong baseline to decide whether to answer a question or abstain \cite{hendrycks2016baseline, lakshminarayanan2017simple, varshney2022investigating}. However, if Model \textit{A} has higher AUC than Model \textit{B}, can we always say that \textit{A} is better at selective answering and thus better suited for safety critical applications than \textit{B}?\\

We experiment across 10 models-- ranging from bag-of-words models to pre-trained transformers-- and find that a model having higher AUC does not necessarily have a higher selective answering capability. We adversarially attack AUC and find several limitations. These limitations prevent AUC from evaluating the efficacy of models in safety critical applications.\\






\begin{table*}[t]
    \centering
    \begin{tabular}{|p{0.85in}|p{0.85in}|p{0.85in}|p{0.85in}|p{0.85in}|p{0.85in}|}
    \hline
    \includegraphics[width=0.85in]{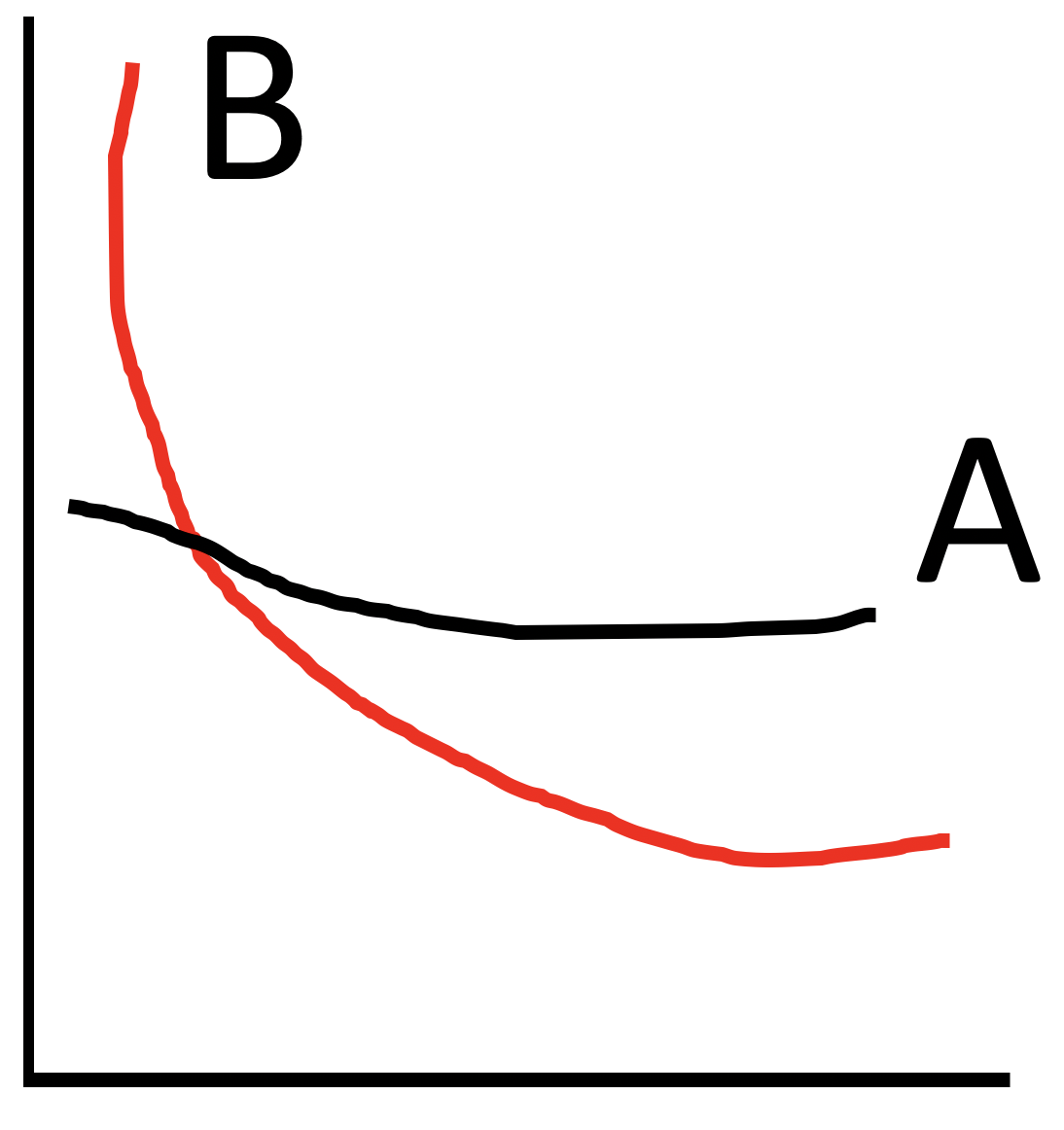}
   &\includegraphics[width=0.85in]{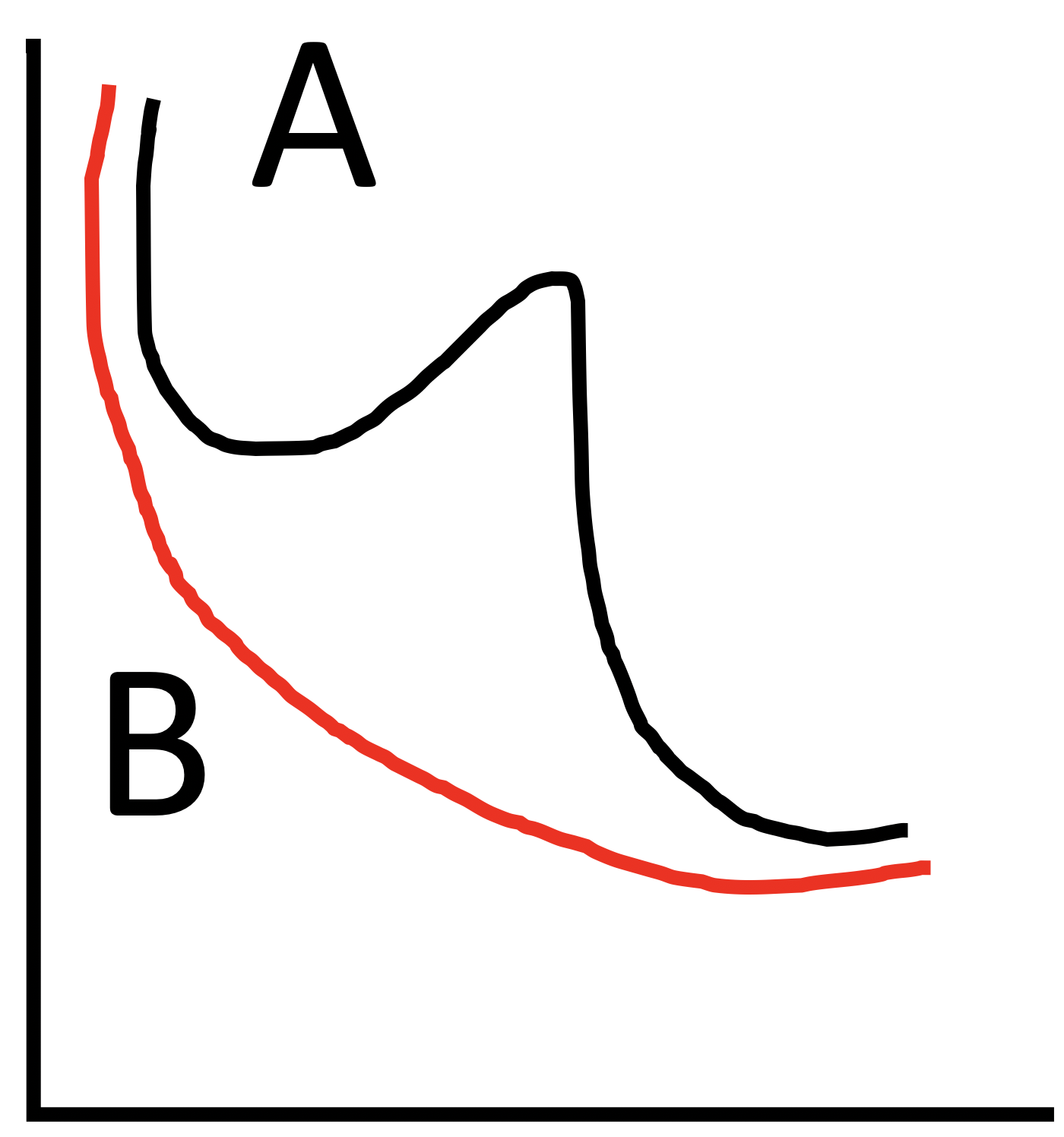}
   &\includegraphics[width=0.85in]{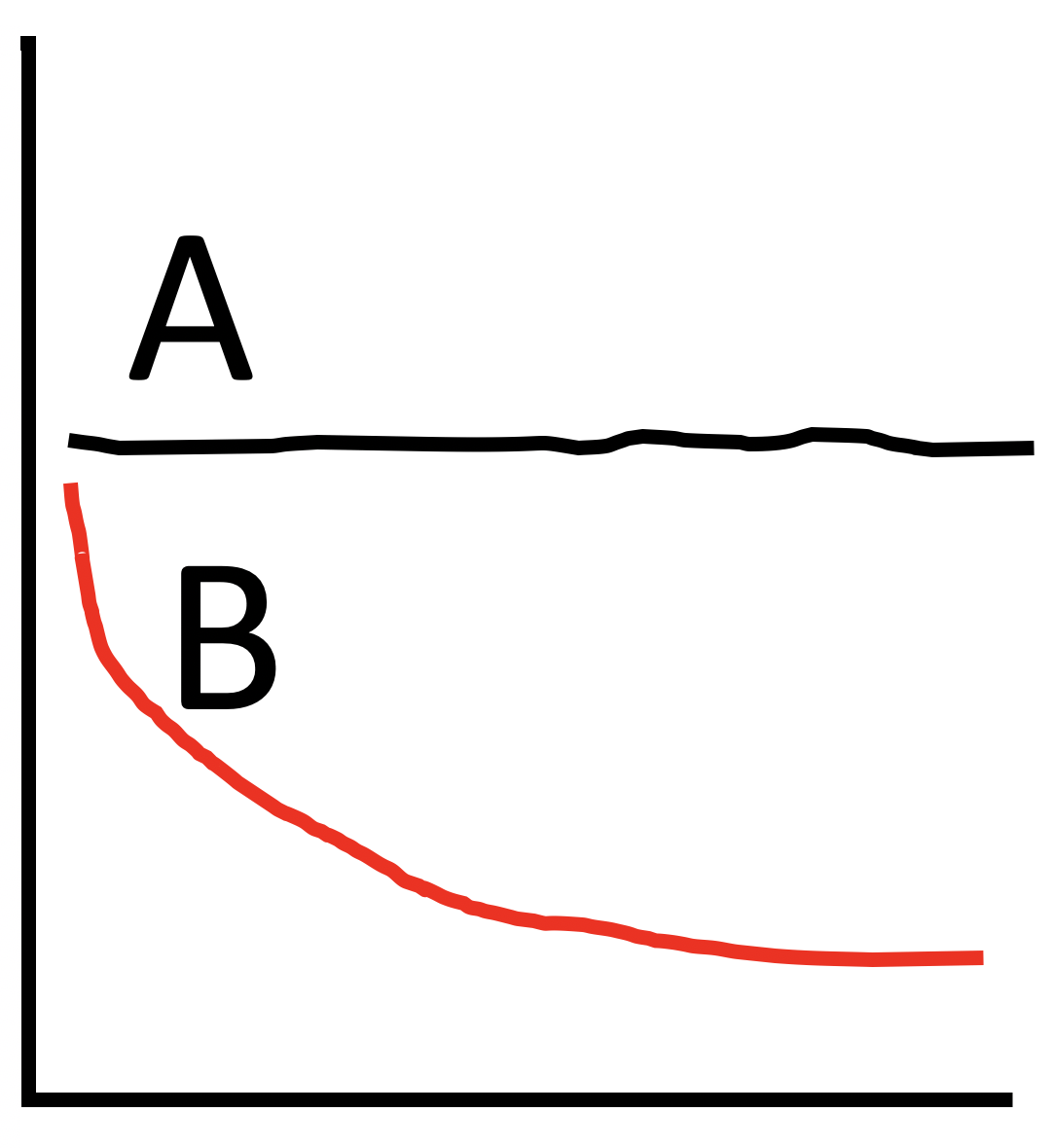}
   &\includegraphics[width=0.85in]{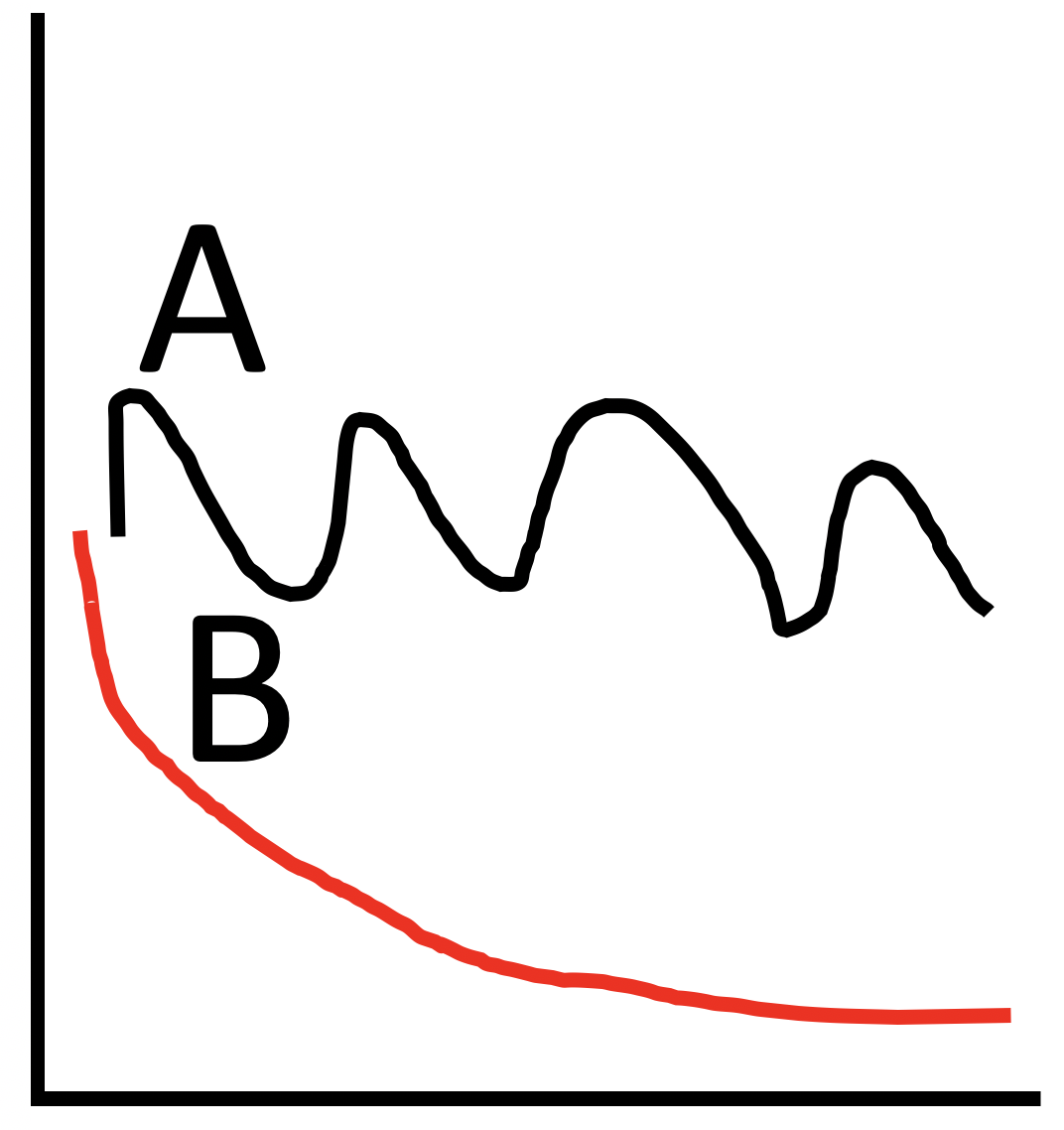}
   &\includegraphics[width=0.85in]{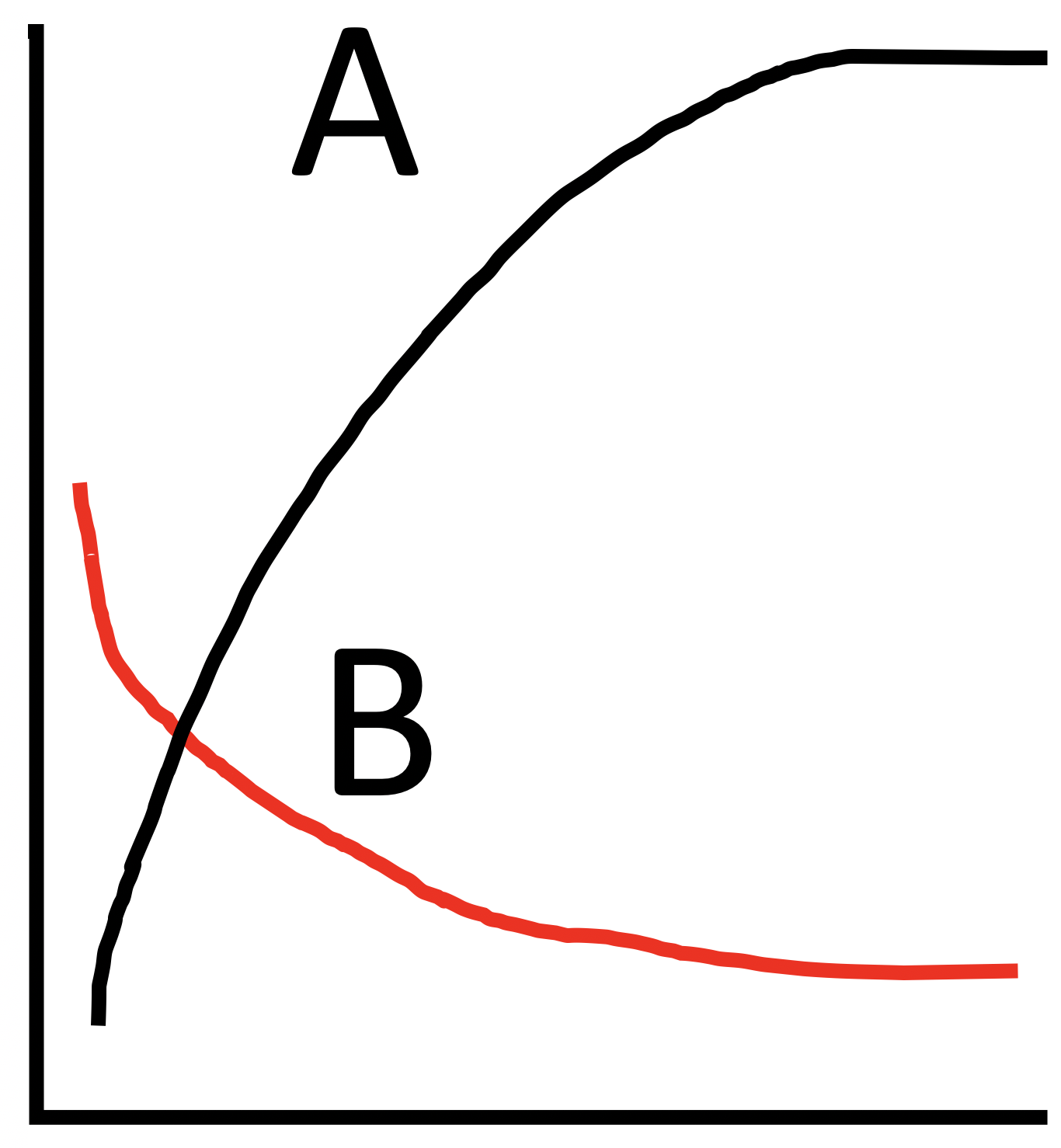}
   &\includegraphics[width=0.85in]{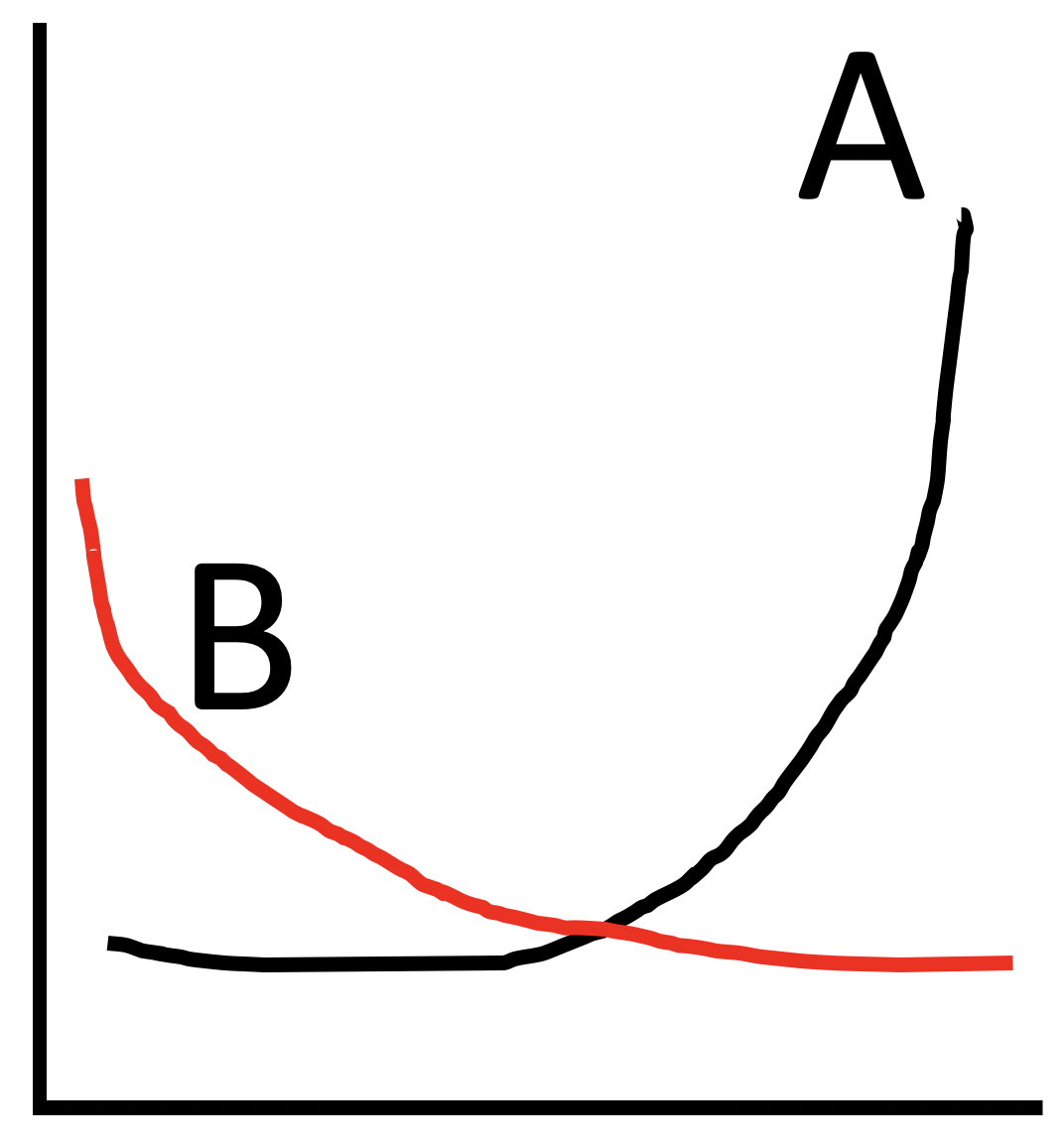}\\\hline
    \end{tabular}
    \caption{In each case, AUC ranks Model A as having higher performance than Model B. However, B has higher selective answering capability, as explained in Section \ref{advat}. X axis: Coverage and Y axis: Accuracy.}
    \label{tab:1}
\end{table*}

In pursuit of fixing the limitations, we propose an evaluation metric-- \textit{`DiSCA Score'} i.e. \textit{Deployment in Safety Critical Applications.} Disaster management is a high-impact and regularly occurring safety critical application. Deployment of Machine Learning systems in disaster management will often be through hand-held devices, such as smartphones, hence requiring light-weight models. Subsequently, we incorporate computation requirements in \textit{`DiSCA Score'}, and propose \textit{`DiDMA Score'} i.e. \textit{Deployment in Disaster Management Applications}. \\ 

NLP applications in disaster management often involve interaction with users ~\cite{phengsuwan2021use, mishra2022towards} whose questions may diverge from a training set, due to rich variations in natural language. We therefore further add the evaluation of abstaining capability on Out-of-Distribution (OOD) datasets as part of the metric and propose \textit{`NiDMA Score'} i.e. \textit{NLP in Disaster Management Applications}. Summarily, (i) \textit{NiDMA} covers interactive NLP applications, (ii) \textit{DiDMA} is specifically tailored to disaster management but can involve non-interactive NLP applications where OOD data is not frequent, and (iii)  \textit{DiSCA} can be used in any safety critical domain.\\

Our analysis across ten models sheds light on the  strengths and weaknesses of various language models. For example, we observe that newer and larger pre-trained models are not necessarily better in performing selective answering. We also observe that model ranking based on the accuracy metric does not match with their ranking based on selective answering capability, similar to the observations by ~\citet{mishra2021robust}. We hope our insights will bring more attention to developing better models and evaluation metrics for safety-critical applications.

\section{Investigating AUC}

\textbf{Experiment Details:}
Our experiments span over ten different models-- Bag-of-Words Sum (BOW-SUM) \cite{harris1954distributional}, Word2Vec Sum (W2V SUM) \cite{mikolov2013distributed}, GloVe Sum (GLOVE SUM) \cite{pennington2014glove}, Word2Vec CNN (W2V CNN) \cite{lecun1995convolutional}, GloVe CNN (GLOVE CNN), Word2Vec LSTM (W2V LSTM) \cite{hochreiter1997long}, GloVe LSTM (GLOVE LSTM),  BERT Base (BERT BASE) \cite{devlin2018bert}, BERT Large (BERT LARGE) with GELU \cite{hendrycks2016gaussian} and RoBERTa Large ((ROBERTA LARGE) \cite{liu2019roberta}, in line with recent works \cite{hendrycks2020pretrained, mishra2020our, mishra2021robust}. We analyze these models over two movie review datasets-- (i) SST-2 \cite{socher2013recursive} that contains short expert movie reviews and (ii) IMDb \cite{maas2011learning} which consists of full-length inexpert movie reviews.
We train models on IMDb and evaluate on both SST-2 and IMDb. Our intuition behind this is to ensure both IID (IMDb test set) and OOD (SST-2 test set) evaluation.

\subsection{Adversarial Attack on AUC:}
\label{advat}

\textbf{AUC Tail:} Consider  a case where \textit{A} has higher overall AUC and lower accuracy than \textit{B}, in regions of higher accuracy and lower coverage (Table \ref{tab:1}, Case 1). Then, \textit{B} is better because safety critical applications have a lower tolerance for incorrect answering and so most often will operate in the region of higher accuracy.  This is seen in the case of the AUCs of BERT-BASE (\textit{A}) and BERT-LARGE (\textit{B}) for the SST-2 dataset \footnote{\label{note2}See Supplementary Material:DiSCA, NiDMA for more details}. \\

\textbf{Curve Fluctuations: }Another case is when accuracy does not vary in a monotonically decreasing fashion. Even though the model has higher AUC, a non-monotonically decreasing curve shape shows that confidence and correctness of answer are not correlated, making the corresponding model comparatively undesirable (Table \ref{tab:1}, Cases 2-6)-- this is seen across all models over both datasets, especially at regions of low coverage. The fluctuations for the OOD dataset are more frequent and have higher magnitude for most models \textsuperscript{\ref{note2}}. \\

\textbf{Plateau Formation:} The range of maximum softmax probability values that models associate with predictions varies. For example we see that LSTM models have a wide maxprob range, while transformer models (BERT-LARGE, ROBERTA-LARGE) answer all questions with high maxprob values. In the latter case, model accuracy stays above values of ~ 90\% in regions of high coverage; this limited accuracy range forms a plateau in the AUC curve. This plateau is not indicative of model performance as the maxprob values (of incorrect answers) in this region are high and relatively unvarying compared to other models; it is therefore undesirable. Such a plateau also makes it difficult to decide which portion of the AUC curve to ignore and find an operating point, while deploying in disaster management applications (where the tolerance for incorrect answering is low). Plateau formation is acceptable when models answer with low maxprob, either always or in regions of high coverage, irrespective of the level of accuracy (though the range should be limited). However, this acceptable curve condition is not observed in any of the models examined, over both datasets.\\

\section{Alternative Metrics}
\subsection{DiSCA Score}
Let $a$ be the maxprob value when accuracy first drops below 100\%. Let $b$ be the lower bound maxprob value which when used as a cutoff for answering questions, results in an accuracy that is the worst possible accuracy admissible by the tolerance level of the domain. Let $n$ represent the number of times the slope of the curve is seen to increase with increasing coverage, and $d_{1},d_{2}$ and $c_{1},c_{2}$ respectively represent the accuracy and maxprob values at the two points where this increase in slope occurs, such that $d_{1} < d_{2}$ and $c_{1} >= c_{2}$. Let $x$, $y$, $z$ represent weights that flexibly define the region of interest depending on application requirements, such that $x+y+z=1$. $b$ is a hyperparameter; we experiment with  a range of $b$ values, where the worst possible accuracy varies from 100\% to 50\%. The \textit{DiSCA Score} is defined as: 
\begin{equation}
    DiSCA= \frac{x}{a} + \frac{y}{b} - \frac{z \cdot \sum_{i}^{n}(n-i+1)\cdot \frac{(d_{2}-d_{1})}{c}}{\sum_{i}^{n}(n-i+1)}
    \label{eq:1}
\end{equation}
where
\vspace{-7mm}
\begin{equation*}
    c=
    \begin{cases}
      (c_{1}-c_{2}) &\text{if}\ (c_{1}-c_{2})>0.001 \\
      0.001 & \text{otherwise}
    \end{cases}
\end{equation*}
Our definition of $c$ is based on the observation of very low order differences in maxprob values for CNN and transformer models. \\

%
\subsubsection{Observations:}
\begin{figure}[H]
    \centering
    \includegraphics[width=0.5\textwidth]{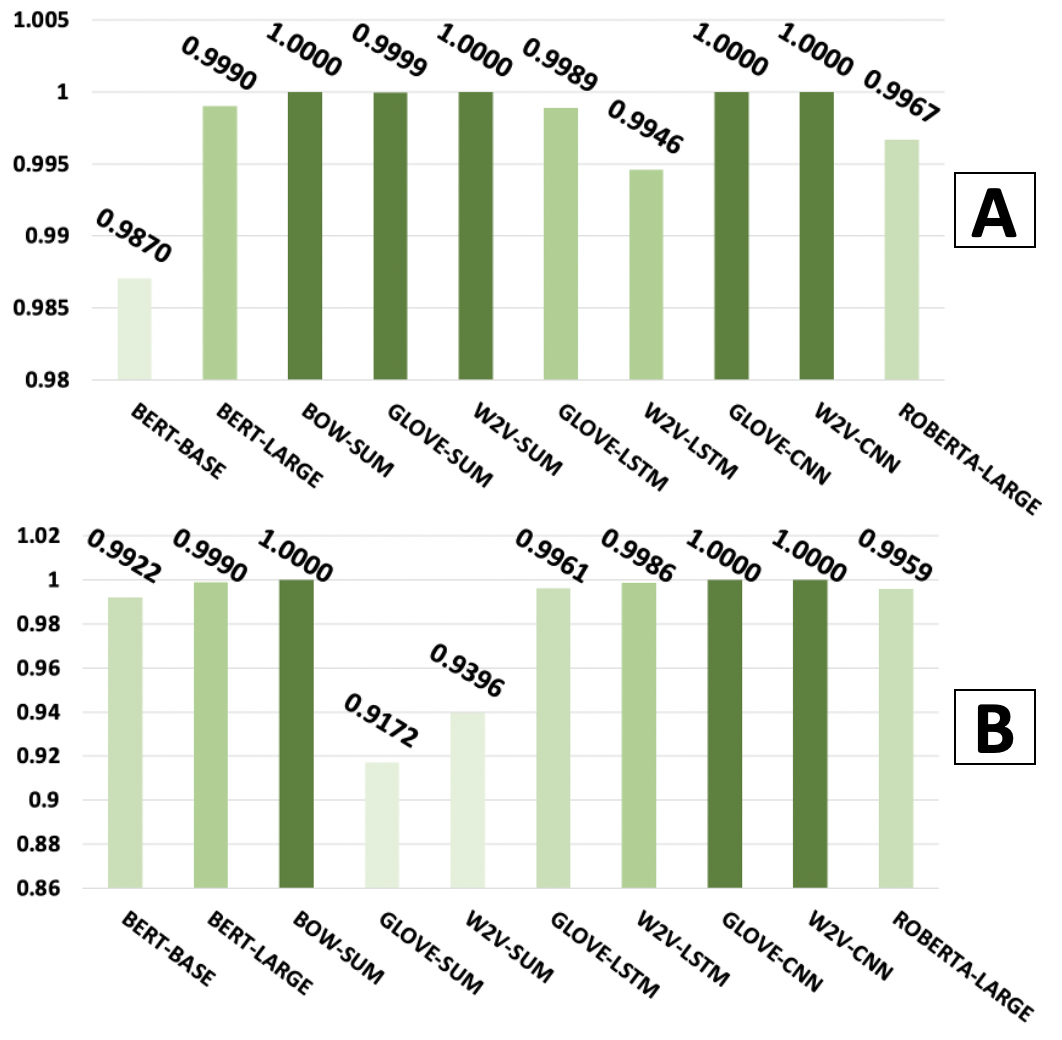}
    \caption{$a$ values for (A)IMDb and (B)SST-2 datasets. IMDb, the IID dataset is used in \textit{DisCA Score}, while SST-2, the OOD dataset is used in  \textit{NiDMA Score}.}
    \label{fig:a}
\end{figure}

\textbf{First Term:}
From Figure \ref{fig:a}(A), we see that the $a$ values are lower for the BERT-BASE, W2V-LSTM and RoBERTA-LARGE models, indicating that the first incorrect classification occurs at lower maxprob values than in other models. Based on equation \ref{eq:1}, BERT-BASE, W2V-LSTM and RoBERTA-LARGE are the top-3 models respectively based on the first term of \textit{DiSCA Score}.\\


\begin{figure*}[t]
    \centering
    \includegraphics[width=0.97\textwidth]{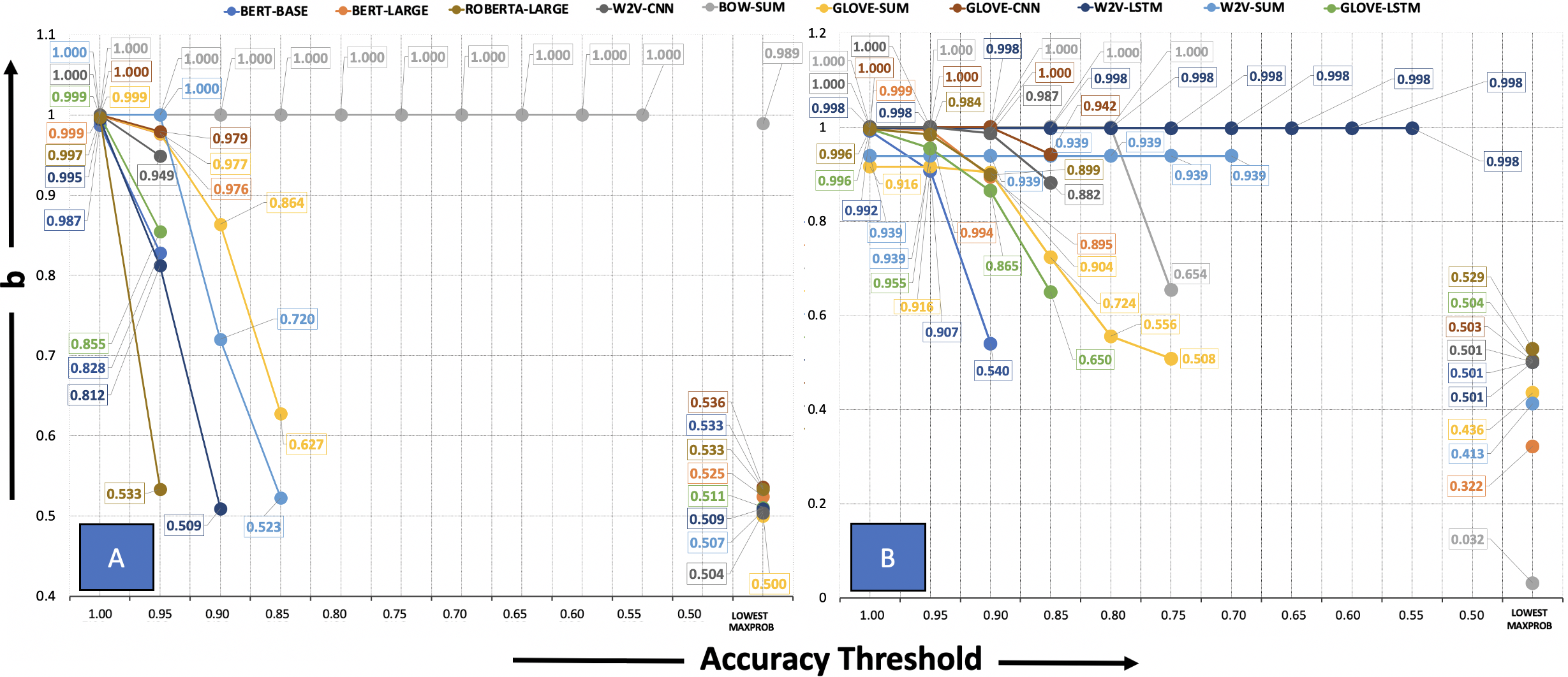}
    \caption{$b$ for (A)IMDb, (B) SST-2 datasets. Values from 1-0.5 on the x-axis represent the accuracy thresholds considered, while `Lowest Maxprob' represents the lowest maxprob value a model assigns at 100\% coverage.}
    \label{fig:bc}
\end{figure*}

\begin{table*}[t]
\centering
\resizebox{0.97\textwidth}{!}{%

\begin{tabular}{
>{\columncolor[HTML]{FFFFFF}}c 
>{\columncolor[HTML]{9AFF99}}c 
>{\columncolor[HTML]{FFFFFF}}c 
>{\columncolor[HTML]{FFFFFF}}c 
>{\columncolor[HTML]{FFFFFF}}c 
>{\columncolor[HTML]{FFFFFF}}c 
>{\columncolor[HTML]{FFFFFF}}c 
>{\columncolor[HTML]{FFFFFF}}c 
>{\columncolor[HTML]{FFFFFF}}c 
>{\columncolor[HTML]{FFFFFF}}c }\hline
\textbf{MODEl} & \cellcolor[HTML]{FFFFFF}\textbf{\textless 0.95} & \textbf{\textless 0.90} & \textbf{\textless 0.85} & \textbf{\textless 0.80} & \textbf{\textless 0.75} & \textbf{\textless 0.70} & \textbf{\textless 0.65} & \textbf{\textless 0.60} & \textbf{\textless 0.55} \\\hline
\textbf{BERT-BASE} & 0.69 & 0.69 & 0.69 & 0.69 & 0.69 & 0.69 & 0.69 & 0.69 & 0.69 \\
\textbf{W2V-LSTM} & 0.63 & \cellcolor[HTML]{9AFF99}0.65 & 0.65 & 0.65 & 0.65 & 0.65 & 0.65 & 0.65 & 0.65 \\
\textbf{GLOVE-CNN} & 0.54 & 0.54 & 0.54 & 0.54 & 0.54 & 0.54 & 0.54 & 0.54 & 0.54 \\
\textbf{GLOVE-SUM} & 0.52 & \cellcolor[HTML]{9AFF99}0.54 & \cellcolor[HTML]{9AFF99}0.56 & 0.56 & 0.56 & 0.56 & 0.56 & 0.56 & 0.56 \\
\textbf{BERT-LARGE} & 0.48 & 0.48 & 0.48 & 0.48 & 0.48 & 0.48 & 0.48 & 0.48 & 0.48 \\
\textbf{ROBERTA-LARGE} & 0.45 & 0.45 & 0.45 & 0.45 & 0.45 & 0.45 & 0.45 & 0.45 & 0.45 \\
\textbf{GLOVE-LSTM} & 0.37 & 0.37 & 0.37 & 0.37 & 0.37 & 0.37 & 0.37 & 0.37 & 0.37 \\
\textbf{W2V-SUM} & 0.35 & \cellcolor[HTML]{9AFF99}0.37 & \cellcolor[HTML]{9AFF99}0.39 & 0.39 & 0.39 & 0.39 & 0.39 & 0.39 & 0.39 \\
\textbf{W2V-CNN} & 0.36 & 0.36 & 0.36 & 0.36 & 0.36 & 0.36 & 0.36 & 0.36 & 0.36 \\
\textbf{BOW-SUM} & \cellcolor[HTML]{FFFFC7}-6.71 & \cellcolor[HTML]{FFFFC7}-6.71 & \cellcolor[HTML]{FFFFC7}-6.71 & \cellcolor[HTML]{FFFFC7}-6.71 & \cellcolor[HTML]{FFFFC7}-6.71 & \cellcolor[HTML]{FFFFC7}-6.71 & \cellcolor[HTML]{FFFFC7}-6.71 & \cellcolor[HTML]{FFFFC7}-6.71 & \cellcolor[HTML]{9AFF99}-6.71\\
\bottomrule
\end{tabular}


\begin{tabular}{
>{\columncolor[HTML]{FFFFFF}}c 
>{\columncolor[HTML]{9AFF99}}c 
>{\columncolor[HTML]{FFFFFF}}c 
>{\columncolor[HTML]{FFFFFF}}c 
>{\columncolor[HTML]{FFFFFF}}c 
>{\columncolor[HTML]{FFFFFF}}c 
>{\columncolor[HTML]{FFFFFF}}c 
>{\columncolor[HTML]{FFFFFF}}c 
>{\columncolor[HTML]{FFFFFF}}c 
>{\columncolor[HTML]{FFFFFF}}c }\hline
\textbf{MODEL} & \cellcolor[HTML]{FFFFFF}\textbf{\textless 0.95} & \textbf{\textless 0.90} & \textbf{\textless 0.85} & \textbf{\textless 0.80} & \textbf{\textless 0.75} & \textbf{\textless 0.70} & \textbf{\textless 0.65} & \textbf{\textless 0.60} & \textbf{\textless 0.55} \\\hline
\textbf{W2V-SUM} & 0.71 & 0.71 & 0.71 & 0.71 & 0.71 & 0.71 & 0.71 & 0.71 & 0.71 \\
\textbf{GLOVE-LSTM} & 0.65 & \cellcolor[HTML]{9AFF99}0.67 & 0.70 & 0.70 & 0.70 & 0.70 & 0.70 & 0.70 & 0.70 \\
\textbf{GLOVE-SUM} & 0.60 & 0.61 & 0.63 & 0.65 & 0.68 & 0.68 & 0.68 & 0.68 & 0.68 \\
\textbf{BERT-LARGE} & 0.60 & \cellcolor[HTML]{9AFF99}0.62 & \cellcolor[HTML]{9AFF99}0.62 & 0.62 & 0.62 & 0.62 & 0.62 & 0.62 & 0.62 \\
\textbf{BERT-BASE} & 0.43 & 0.45 & 0.45 & 0.45 & 0.45 & 0.45 & 0.45 & 0.45 & 0.45 \\
\textbf{ROBERTA-LARGE} & 0.22 & 0.24 & 0.24 & 0.24 & 0.24 & 0.24 & 0.24 & 0.24 & 0.24 \\
\textbf{W2V-CNN} & -0.68 & -0.66 & -0.64 & -0.64 & -0.64 & -0.64 & -0.64 & -0.64 & -0.64 \\
\textbf{BOW-SUM} & -0.71 & \cellcolor[HTML]{9AFF99}-0.69 & \cellcolor[HTML]{9AFF99}-0.68 & -0.66 & -0.63 & -0.63 & -0.63 & -0.63 & -0.63 \\
\textbf{GLOVE-CNN} & -2.43 & -2.43 & -2.42 & -2.42 & -2.42 & -2.42 & -2.42 & -2.42 & -2.42 \\
\textbf{W2V-LSTM} & \cellcolor[HTML]{FFFFC7}-3.53 & \cellcolor[HTML]{FFFFC7}-3.53 & \cellcolor[HTML]{FFFFC7}-3.53 & \cellcolor[HTML]{FFFFC7}-3.53 & \cellcolor[HTML]{FFFFC7}-3.53 & \cellcolor[HTML]{FFFFC7}-3.53 & \cellcolor[HTML]{FFFFC7}-3.53 & \cellcolor[HTML]{FFFFC7}-3.53 & \cellcolor[HTML]{9AFF99}-3.53\\
\bottomrule
\end{tabular}

}
\caption{(a) LHS: \textit{DiSCA (IID)}-- (IMDb), (b) RHS: \textit{DiSCA (OOD)}-- SST-2; $x$=$y$=$z$=0.33. The column header indicates the accuracy threshold used to calculate $b$. Green: Model accuracy falls below the threshold indicated by the column head, but is higher than the next column threshold. Yellow: If the latter is false. White: Model's overall accuracy is higher than the threshold of that column, and therefore never falls below that threshold. }
\label{tab:2}
\end{table*}


\textbf{Second Term:} Figure \ref{fig:bc}(A) illustrates the $b$ values of various models, over a range of worst possible accuracies. In RoBERTA-LARGE, when the accuracy drops below 95\%, the maxprob value is 0.53; it is above 0.8 for other models. This shows that RoBERTA-LARGE is better than other models for the 90-95\% accuracy bin. GLOVE-SUM is found to be relatively better overall, than other models, as its maxprob is a better indicator of accuracy (seen from sharper decrease in the figure). BOW-SUM is the worst model as a significant amount of samples with the highest confidence values are classified incorrectly, causing $b$ for all accuracy bins  to be extremely high and relatively uniform.\\

\textbf{Third Term:} The number of  times accuracy increases with increase in coverage is highest for the BOW-SUM model, making it the worst model (Figure \ref{fig:z}). Word averaging models (W2V-SUM, BOW-SUM, and GLOVE-SUM) are seen to have a higher number of fluctuations on average; other models have near-zero fluctuations, mostly occurring at the highest maxprob samples \textsuperscript{\ref{note2}}. The magnitude-number ratio of fluctuations in RoBERTA-LARGE is high, in comparison to other models.\\

\textbf{Overall Ranking:}
In Table \ref{tab:2}(a), BERT-BASE is ranked highest, as it has no fluctuation penalty and also has a lower maxprob value when accuracy first drops below 100\% ($a$). Since the magnitude-number fluctuation ratio of BERT-LARGE and RoBERTA-LARGE are higher than that seen for W2V-LSTM, GLOVE-CNN, and GLOVE-SUM, the former are ranked lower. 
Despite GLOVE-SUM having the overall best maxprob-accuracy correlation, its higher fluctuation number lowers the score. BOW-SUM is seen to be the worst (and only negatively scoring model), in line with observations made across all individual terms. \\

\subsection{DiDMA Score}
\begin{equation}
\normalsize
    DiDMA= p*DiSCA+q*Computation\_Score
\end{equation}
where
\begin{equation*}
    Computation\_Score=\frac{1}{Energy}
    \label{eq:2}
\end{equation*}

\textit{DiDMA Score} is obtained by summing weighted \textit{DiSCA Score} ($p$)and \textit{Computation Score} ($q$) based on application requirements (such that $p$+$q$=1). In \textit{Computation Score}, models are scored based on the energy usage. Higher energy usage implies lower computation score. We suggest using equations\footnote{See Supplementary: DiDMA for more details} of a recent work \cite{henderson2020towards} for the energy calculation. Since, energy usage is hardware dependent, we do not calculate the term here; it should be calculated based on the device of deployment in disaster management. 

\subsection{NiDMA Score}
\begin{equation}
    NiDMA= u*DiDMA+v*DiSCA(OOD)
\end{equation}

\begin{figure}[H]
    \centering
    \includegraphics[width=0.48\textwidth]{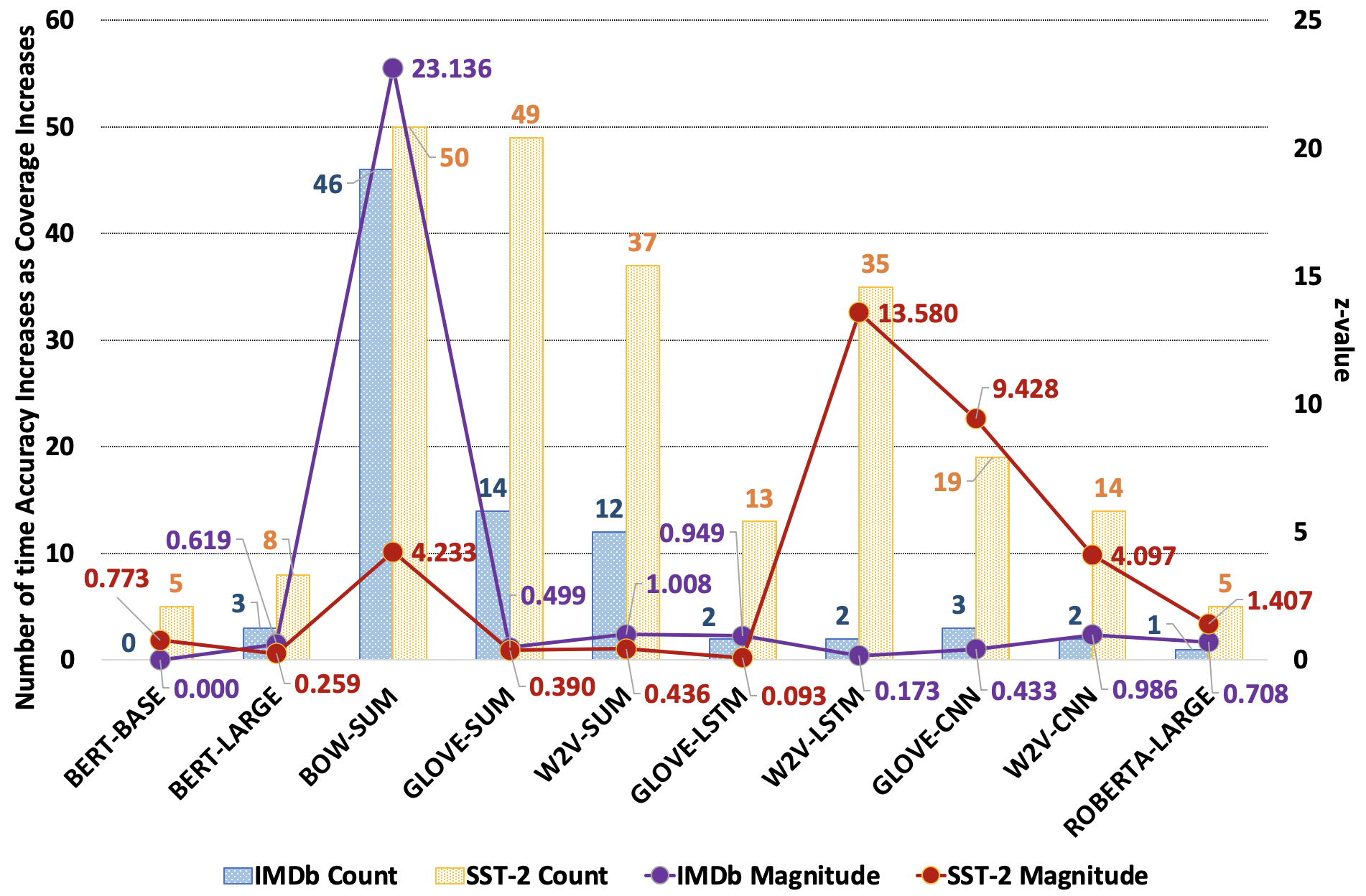}
    \caption{The bars indicate the number of times accuracy is seen to increase as coverage increases and the lines indicate the magnitude of the third term in \ref{eq:1}, over the IMDb and SST-2 datasets.}
    \label{fig:z}
\end{figure}

\textit{NiDMA Score} is obtained by summing weighted \textit{DiDMA Score}($u$) and \textit{DiSCA Score} on OOD data ($v$) based on application requirements (where $u$+$v$=1). Figure \ref{fig:a}(B), \ref{fig:bc}(B) and \ref{fig:z} (SST-2) illustrate the \textit{DiSCA Score} on OOD datasets.\\

From Figure \ref{fig:a}(B), we see that first term based ranking of the \textit{DiSCA(OOD)Score} is not preserved with respect to \textit{DiSCA Score} in Figure \ref{fig:a}(A); $a$ values are lowest for the GLOVE-SUM and W2V-SUM models. We also note that for the $b$ values (Figure \ref{fig:bc}(B)), W2V-LSTM is the worst model, while GLOVE-SUM remains better overall; BERT-BASE and GLOVE-LSTM perform best for the 85-90\% and 80-85\% accuracy bins respectively .\\

\noindent  while other transformer/embeddings models show higher maxprob values ($>$0.8) at higher coverage. In Figure \ref{fig:z}, while there is an increase in the number of accuracy fluctuations, we see that model ranking for the number of times accuracy increases is preserved for most models (except LSTM and CNN). The magnitude decreases for BERT-LARGE, word-averaging, and GLOVE-LSTM. The number-magnitude ratio remains high for RoBERTA-LARGE. We also observe a significant change in ranking from Table \ref{tab:2}(b), though BERT-LARGE and RoBERTA-LARGE have similar ranking; W2V-SUM is found to have best overall performance in terms of \textit{DiSCA (OOD) Score}. \\

\section{Conclusion}
We adversarially attack the AUC metric that is commonly used for evaluating selective answering capability of models. We find that a model having higher AUC is not always better in terms of its selective answering capability, and subsequently its efficacy in safety critical applications. We propose three alternate metrics to fix the limitations in AUCs. We experiment across ten models and get various insights regarding strengths ans weakness of models. We hope our work will encourage the development of better models and evaluation metrics focusing on the safety requirements of users while interacting with machine learning models, that is getting increasingly popular in the instruction paradigm~\cite{mishra2022cross, sanh2021multitask, mishra2021reframing, wei2021finetuned, parmar2022boxbart,ouyang2022training}.

\bibliography{custom}
\bibliographystyle{acl_natbib}
\clearpage
\appendix
\section{DiSCA}
\begin{figure*}[t]
    \centering
    \includegraphics[width=\textwidth]{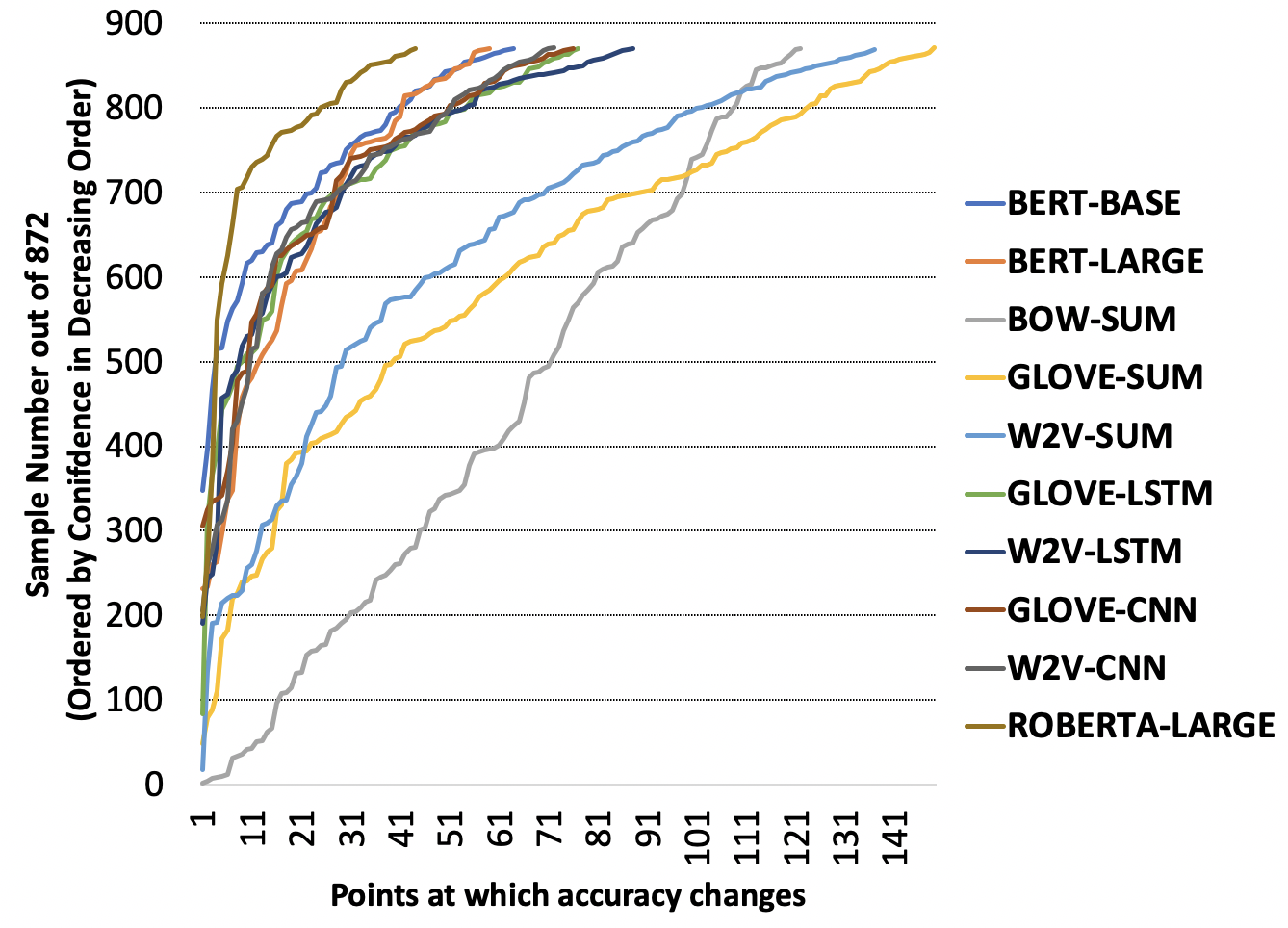}
    \caption{IMDb Samples are ranked in decreasing order of maxprob values for each model. The number of samples classified as per this ordering, at each point the accuracy changes (from a range of 1 to 0.5), is shown.}
    \label{fig:imdb_bn}
\end{figure*}

\begin{figure*}[t]
    \centering
    \includegraphics[width=\textwidth]{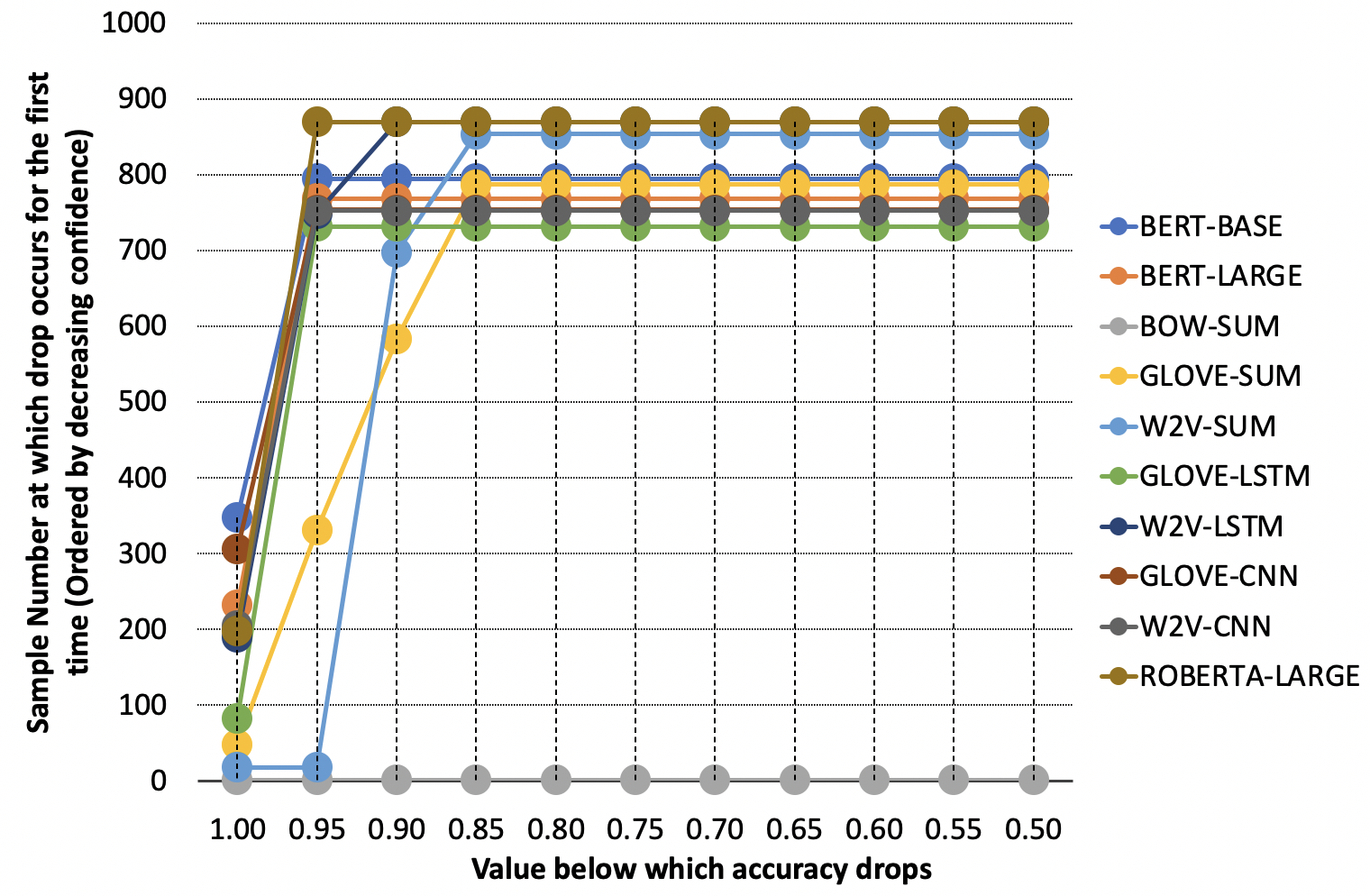}
    \caption{IMDb Samples are ranked in decreasing order of maxprob values for each model. The number of samples classified as per this ordering, at the first point models fail to cross a range of accuracy thresholds (represented on the x-axis), is shown.}
    \label{fig:imdb_bn1}
\end{figure*}

\begin{figure*}[t]
    \centering
    \includegraphics[width=\textwidth]{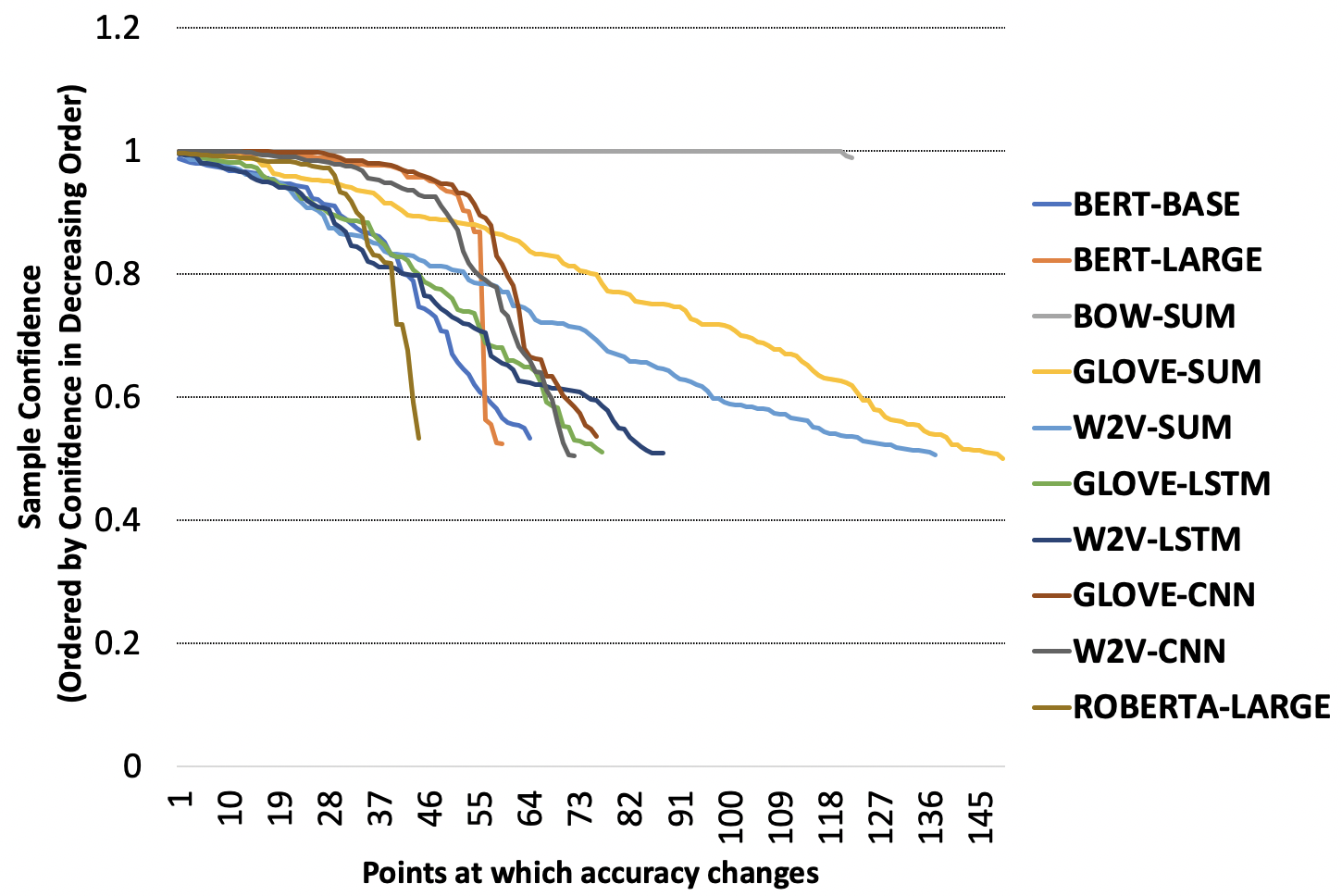}
    \caption{IMDb Samples are ranked in decreasing order of maxprob values for each model. The maxprob value of the latest sample classified as per this ordering, at each point the accuracy changes (from a range of 1 to 0.5), is shown.}
    \label{fig:imdb_bc}
\end{figure*}

\begin{figure*}[t]
    \centering
    \includegraphics[width=\textwidth]{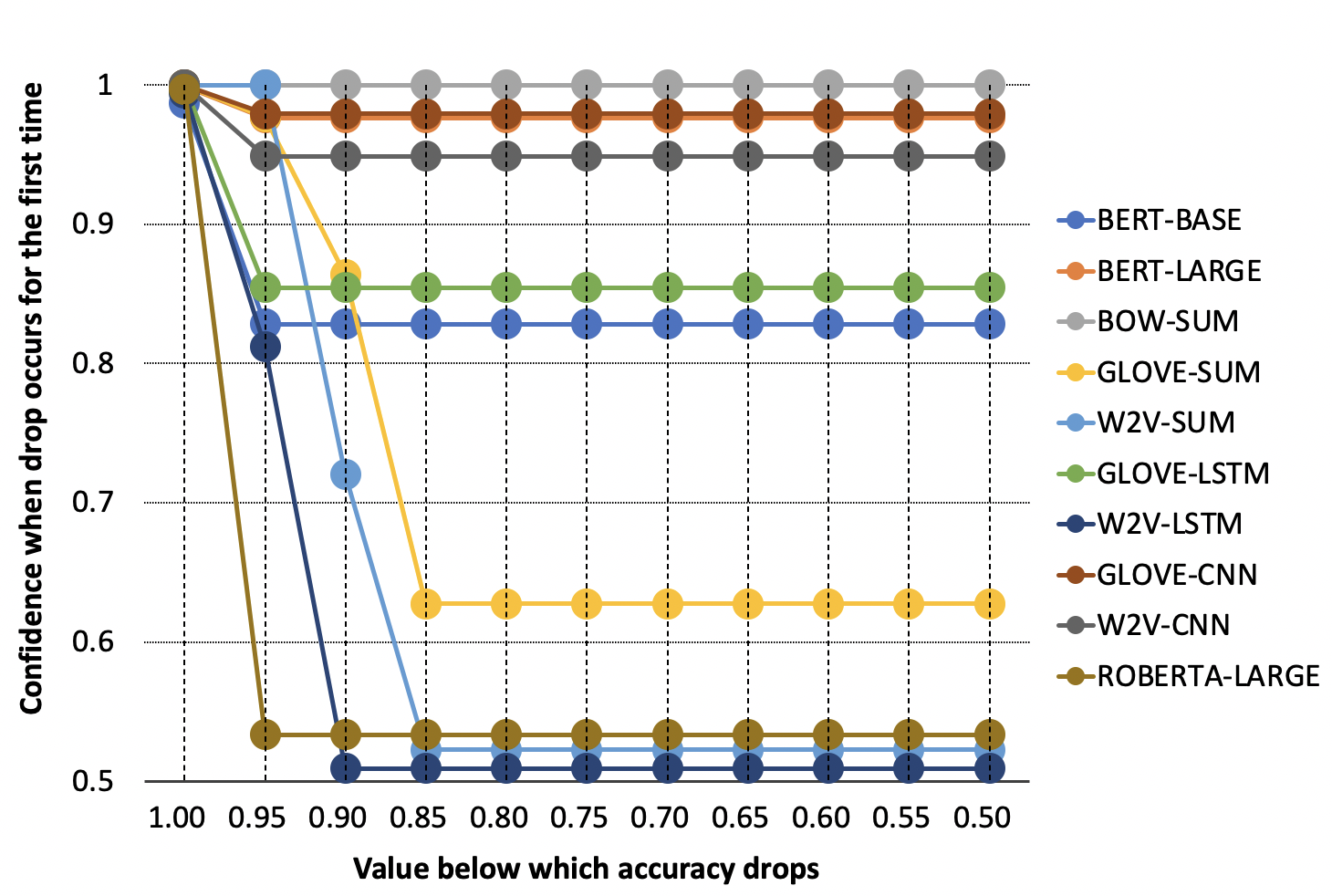}
    \caption{IMDb Samples are ranked in decreasing order of maxprob values for each model. The maxprob of the latest sample classified as per this ordering, at the first point models fail to cross a range of accuracy thresholds (represented on the x-axis), is shown.}
    \label{fig:imdb_bc1}
\end{figure*}

\begin{figure*}[t]
    \centering
    \includegraphics[width=\textwidth]{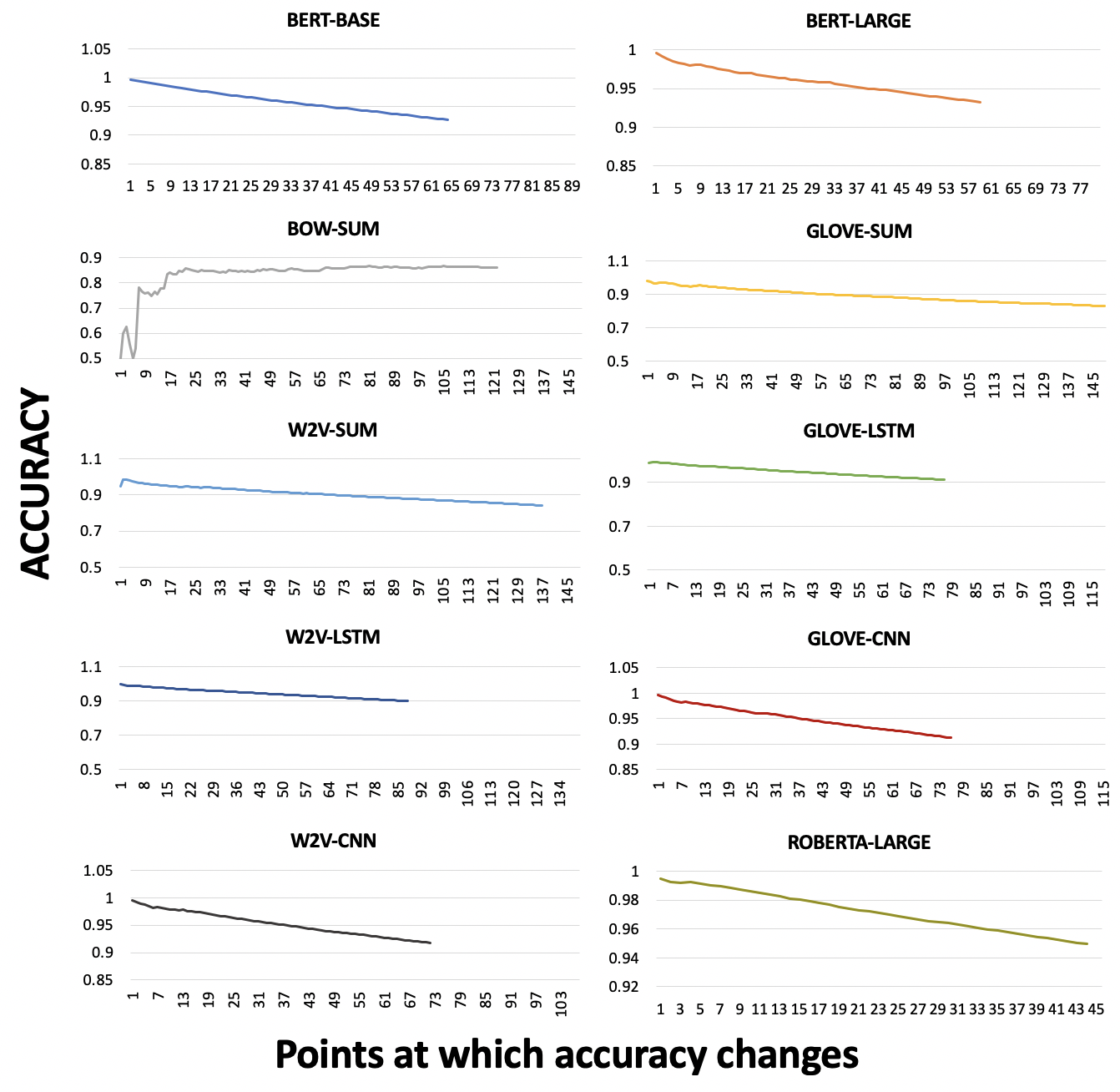}
    \caption{Discrete AUCs of the models are shown, by plotting the accuracy change over increase in coverage (based maxprob thresholding of IMDb samples).}
    \label{fig:imdb_ba}
\end{figure*}

\begin{figure*}[t]
    \centering
    \includegraphics[width=\textwidth]{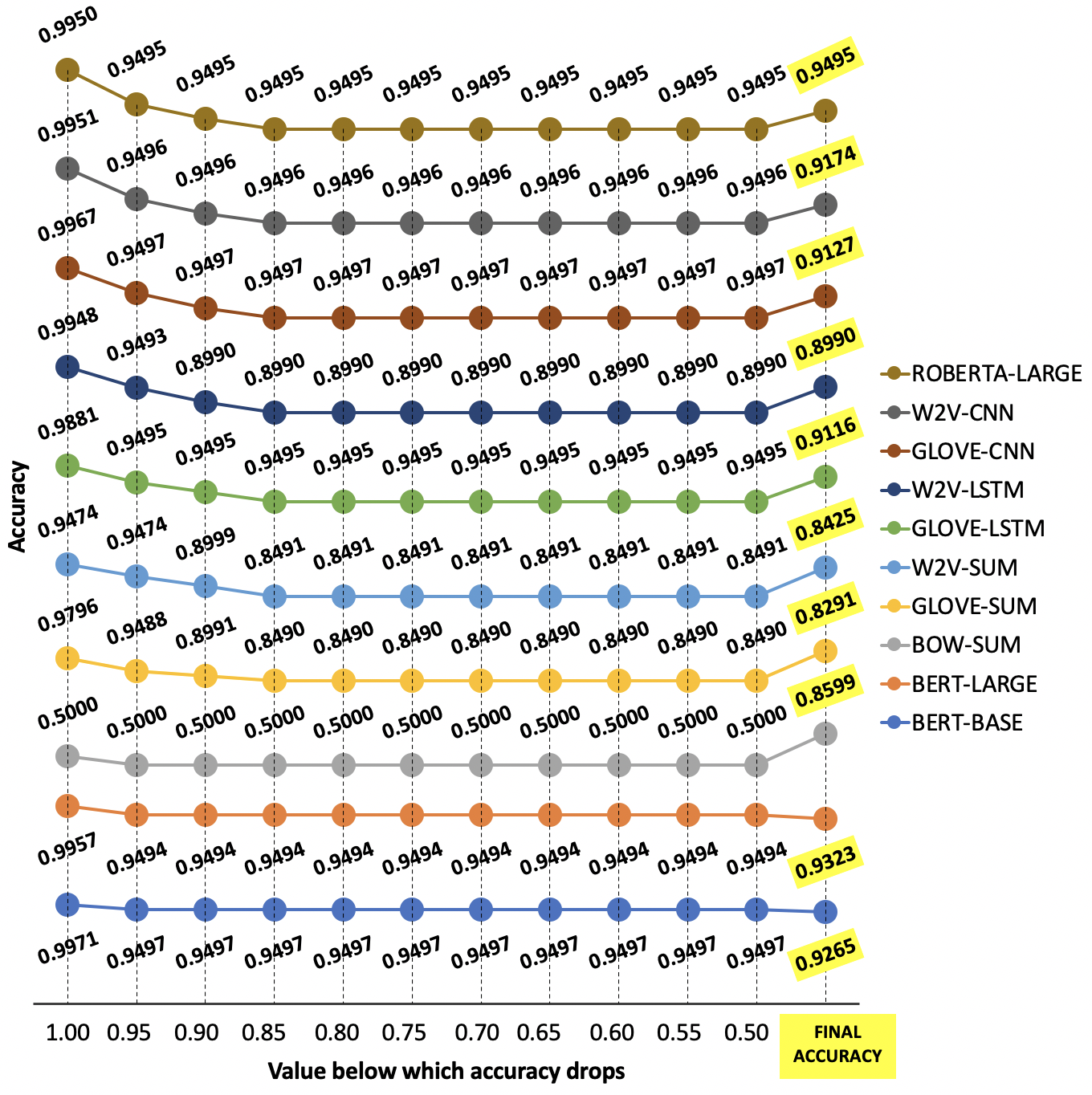}
    \caption{Accuracies at the first point models fail to cross a range of accuracy thresholds (represented on the x-axis), are shown for the  IMDb  dataset.}
    \label{fig:imdb_ba1}
\end{figure*}

Please refer to Figures \ref{fig:imdb_bn},\ref{fig:imdb_bn1},\ref{fig:imdb_bc},\ref{fig:imdb_bc1},\ref{fig:imdb_ba},\ref{fig:imdb_ba1}, \ref{fig:scores1}(A).

\section{DiDMA}
Equation for Computation Score \cite{henderson2020towards}:

\begin{equation}
    e_{total} = PUE \sum_{p}^{} (p_{dram}e_{dram}+p_{cpu}e_{cpu}+p_{gpu}e_{gpu})
\end{equation}

where $p_{resource}$ are the percentages of each system resource used by the attributable processes relative to the total in-use resources and $e_{resource}$ is the energy usage of that resource. The same constant power usage effectiveness (PUE) as \cite{strubell2019energy} is used. This value compensates for excess energy from cooling or heating the data-center.

Alternatively, computation score can be calculated as the ratio of a fixed `optimal' parameter number (based on the hardware used in deployment) to the number of parameters in the model considered. For example, if a deployment device functions effectively when models with up to 1 million parameters are utilized, a model with 2 million parameters will be assigned a score of 0.5 and one with 500,000 parameters will be assigned a score of 2. 

Other efficiency based evaluation in terms of data~\cite{mishra2020we, su2022selective} or model~\cite{min2021neurips, treviso2022efficient} can be incorporated in this metric as part of this parameter in future.

\section{N-DiDMA}
\begin{figure*}[t]
    \centering
    \includegraphics[width=\textwidth]{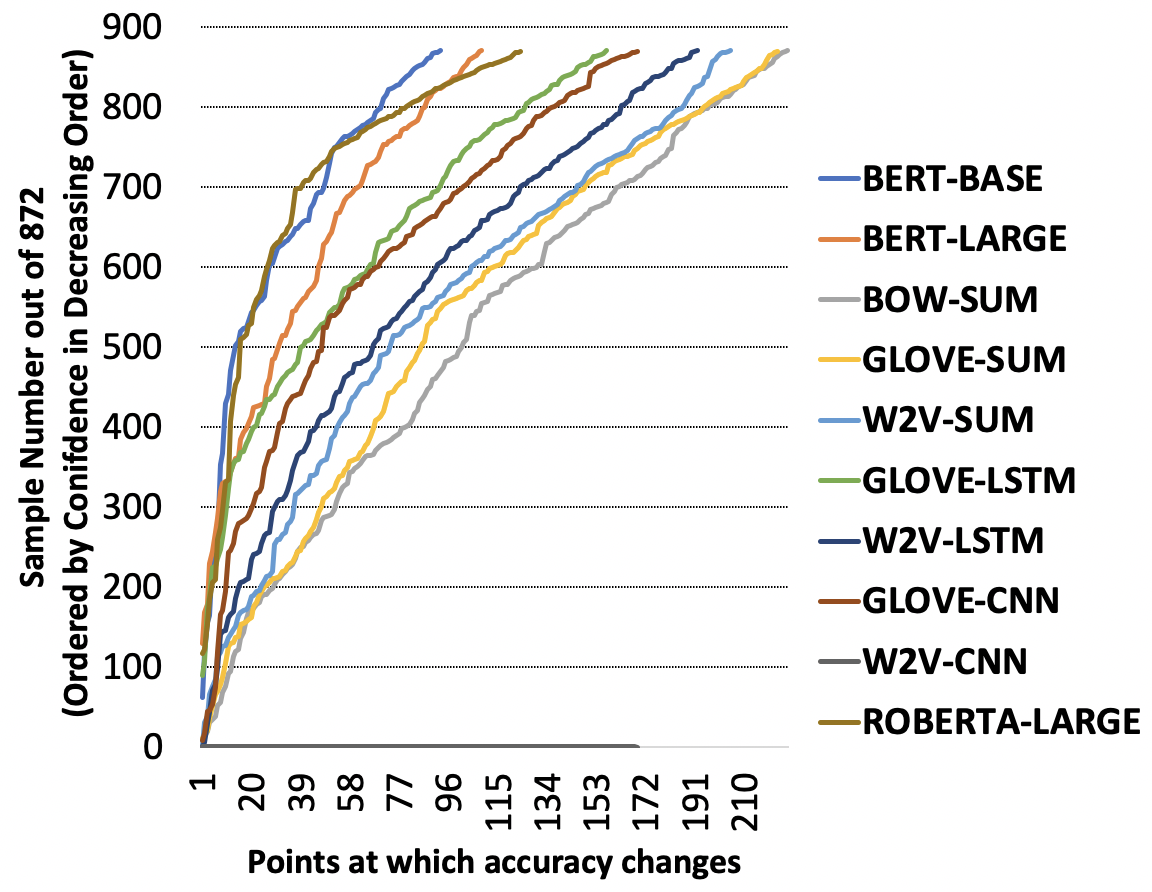}
    \caption{SST-2 Samples are ranked in decreasing order of maxprob values for each model. The number of samples classified as per this ordering, at each point the accuracy changes (from a range of 1 to 0.5), is shown.}
    \label{fig:SST_bn}
\end{figure*}

\begin{figure*}[t]
    \centering
    \includegraphics[width=\textwidth]{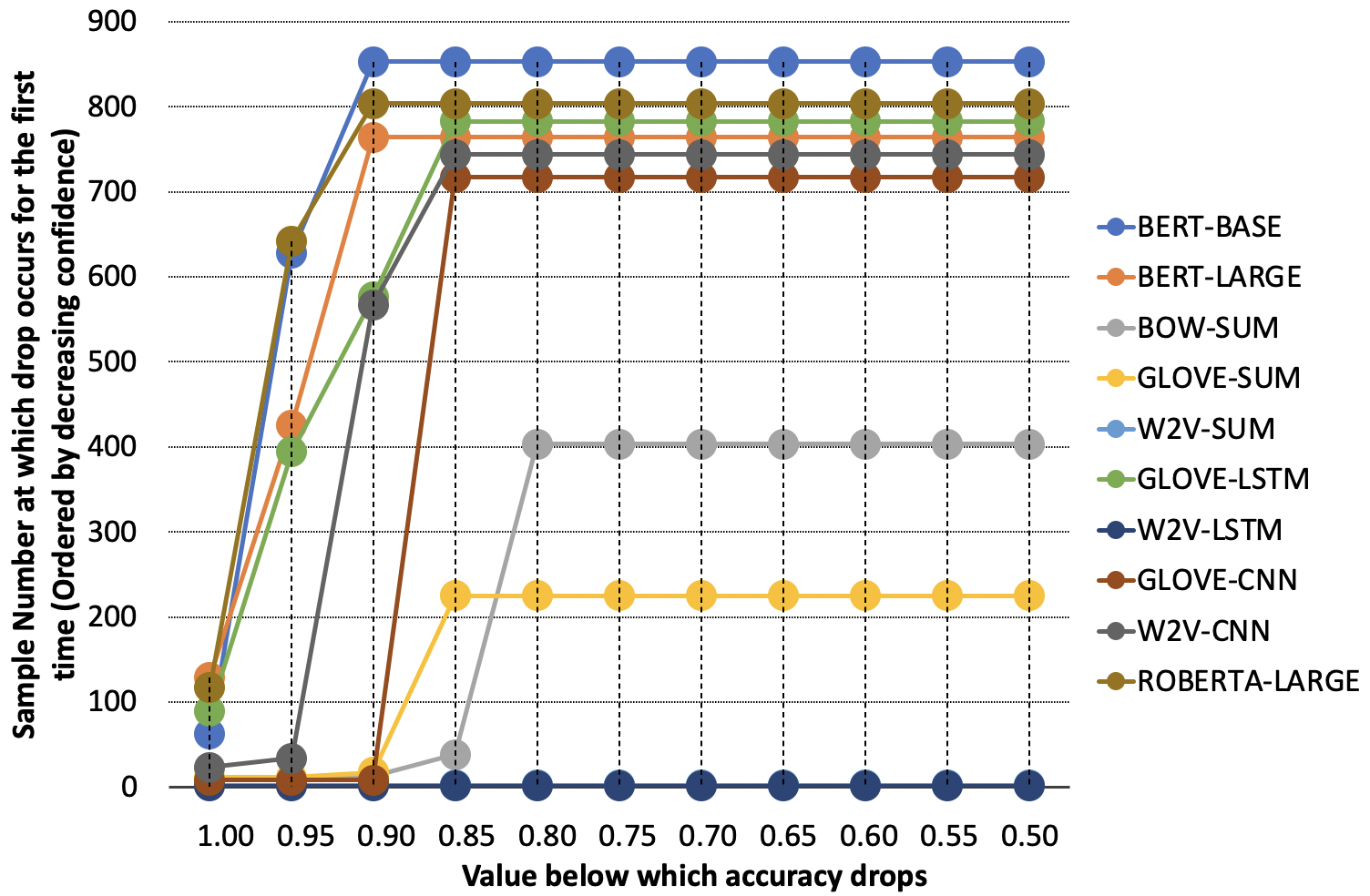}
    \caption{SST-2 Samples are ranked in decreasing order of maxprob values for each model. The number of samples classified as per this ordering, at the first point models fail to cross a range of accuracy thresholds (represented on the x-axis), is shown.}
    \label{fig:SST_bn1}
\end{figure*}

\begin{figure*}[t]
    \centering
    \includegraphics[width=\textwidth]{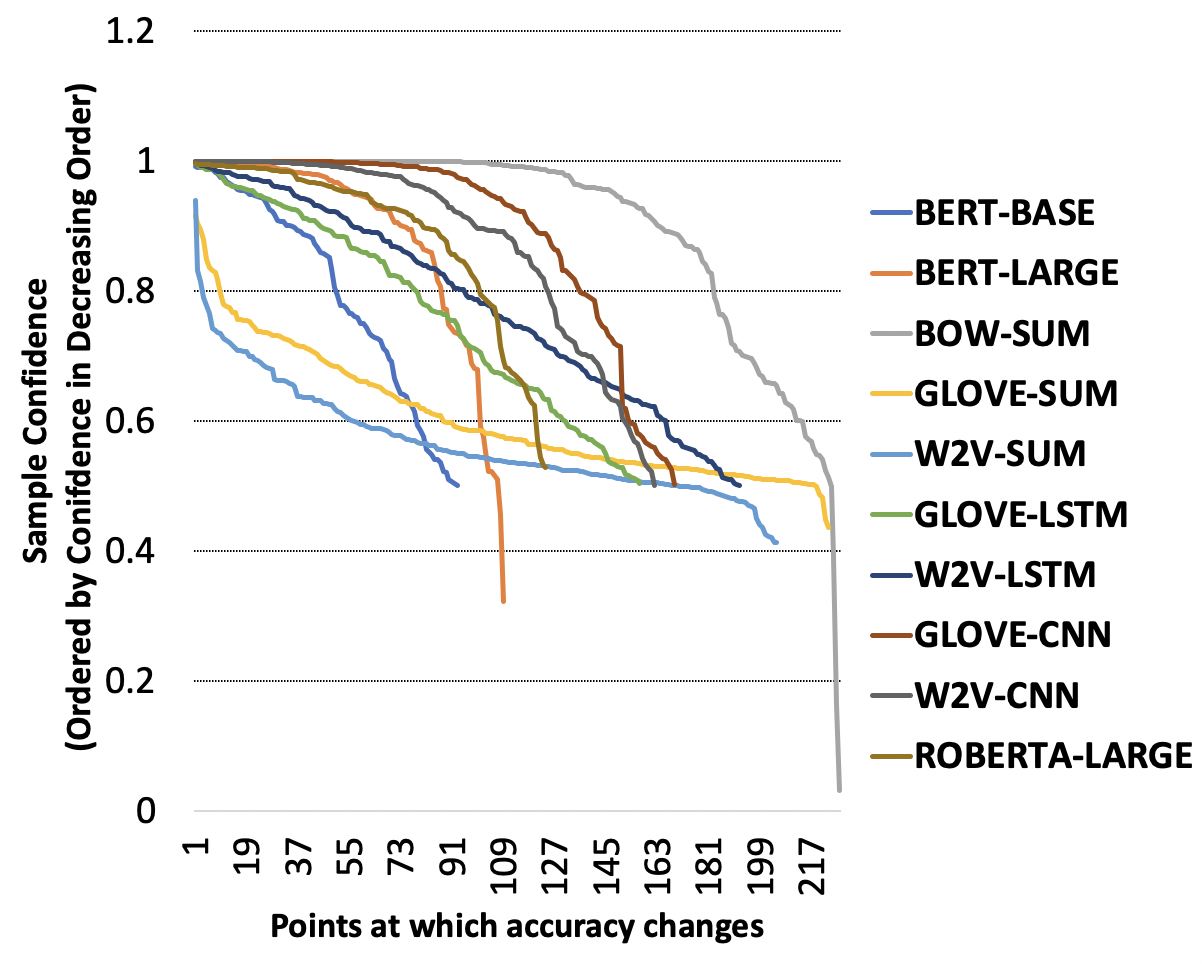}
    \caption{SST-2 Samples are ranked in decreasing order of maxprob values for each model. The maxprob value of the latest sample classified as per this ordering, at each point the accuracy changes (from a range of 1 to 0.5), is shown.}
    \label{fig:SST_bc}
\end{figure*}

\begin{figure*}[t]
    \centering
    \includegraphics[width=\textwidth]{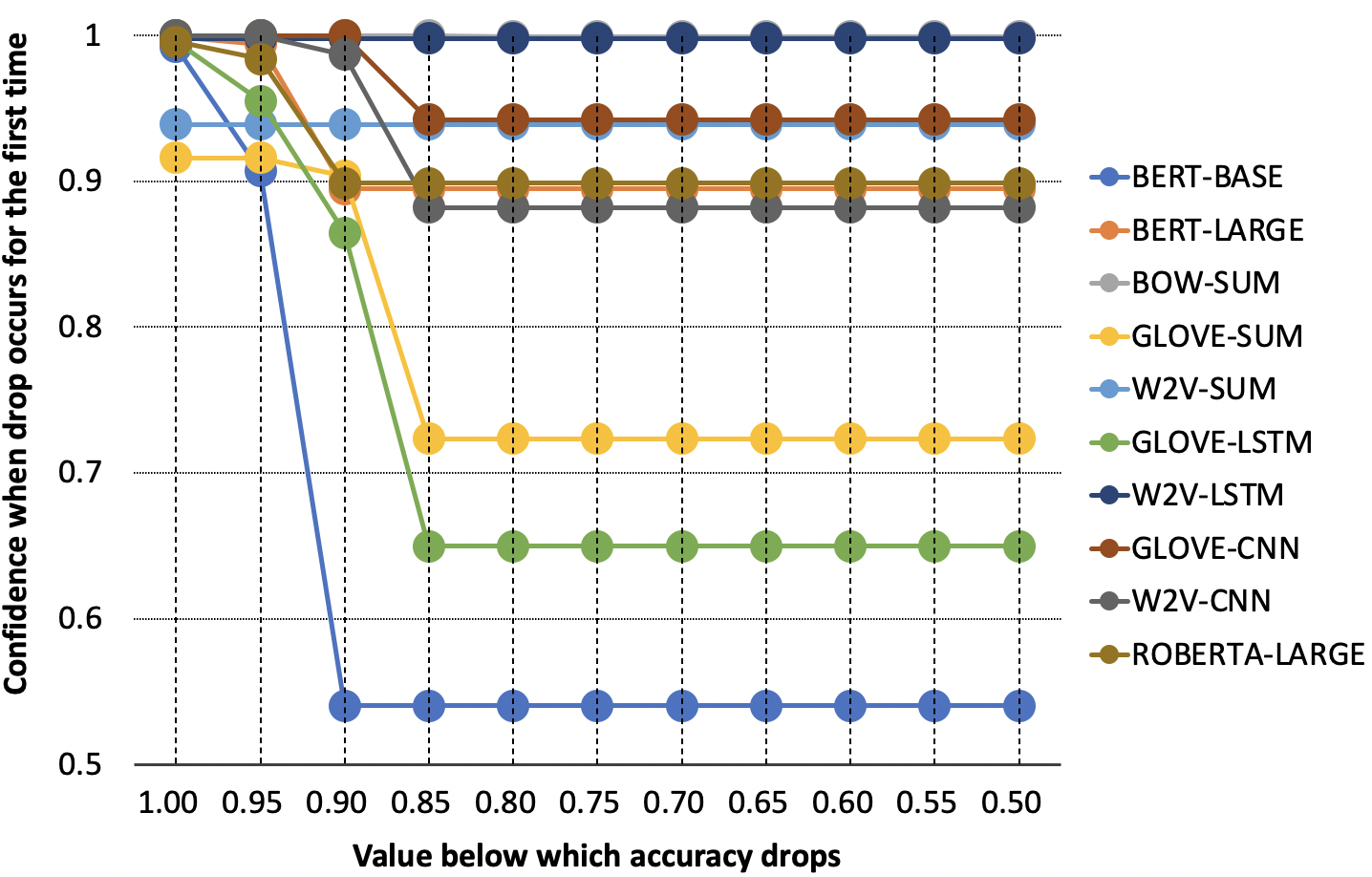}
    \caption{SST-2 Samples are ranked in decreasing order of maxprob values for each model. The maxprob of the latest sample classified as per this ordering, at the first point models fail to cross a range of accuracy thresholds (represented on the x-axis), is shown.}
    \label{fig:SST_bc1}
\end{figure*}

\begin{figure*}[t]
    \centering
    \includegraphics[width=\textwidth]{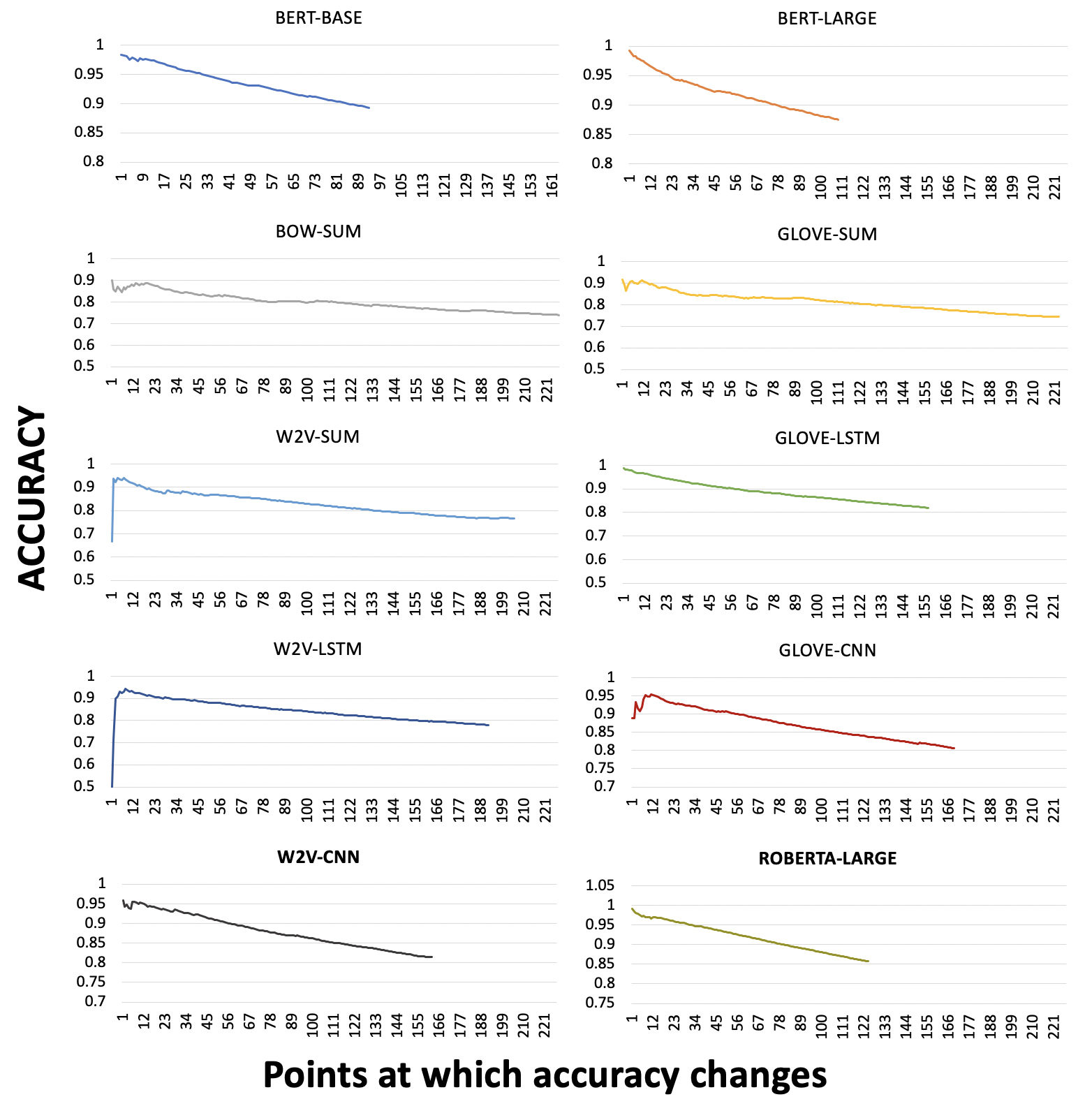}
    \caption{Discrete AUCs of the models are shown, by plotting the accuracy change over increase in coverage (based maxprob thresholding of SST-2 samples).}
    \label{fig:SST_ba}
\end{figure*}

\begin{figure*}[t]
    \centering
    \includegraphics[width=\textwidth]{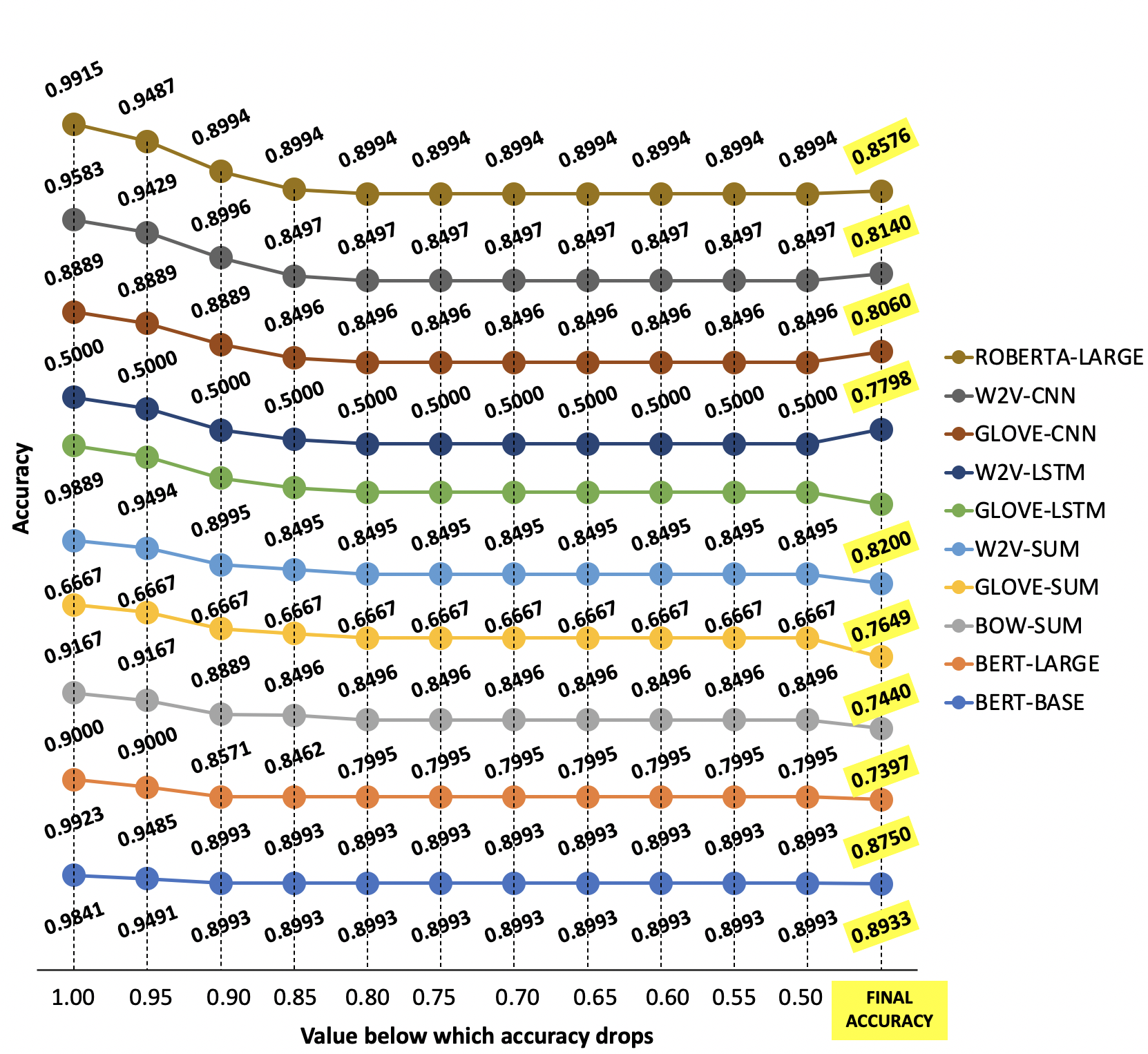}
    \caption{Accuracies at the first point models fail to cross a range of accuracy thresholds (represented on the x-axis), are shown for the  SST-2  dataset.}
    \label{fig:SST_ba1}
\end{figure*}

\begin{figure*}[t]
    \centering
    \includegraphics[width=\textwidth]{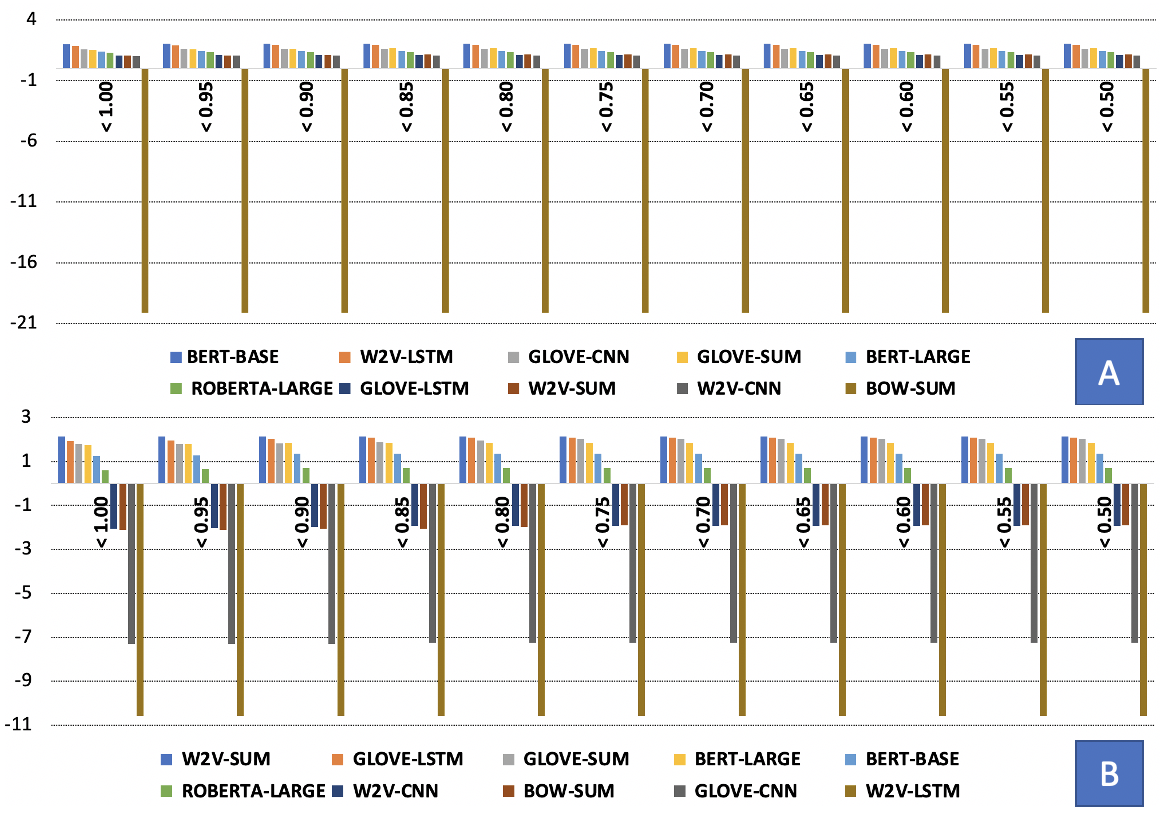}
    \caption{Model ranking based on IMDb (A) and SST-2 (B) performances. The x-axis represents the accuracy threshold considered for $b$.}
    \label{fig:scores1}
\end{figure*}

Please refer to Figures \ref{fig:SST_bn}, \ref{fig:SST_bn1}, \ref{fig:SST_bc}, \ref{fig:SST_bc1}, \ref{fig:SST_ba}, \ref{fig:SST_ba1}, \ref{fig:scores1}(B).

N-DiDMA and DiDMA are narrower in scope as they are specific to NLP and disaster management applications whereas DiSCA can be applied in general in any safety-critical applications, for example the cyber-physical smart grid~\cite{mishra2015generalized, korukonda2016improving, mishra2019enabling}.
\section{Infrastructure Used}
All the experiments were conducted on ”TeslaV100-SXM2-16GB”; CPU cores per node 20; CPU memory
per node: 95,142 MB; CPU memory per core: 4,757 MB. This configuration is not a necessity for these
experiments as we ran our operations with NVIDIA Quadro RTX 4000 as well with lesser memory.



\end{document}